\documentclass[pdflatex,sn-nature,oneside]{sn-jnl}%

\usepackage{graphicx}%
\newcommand{\panel}[1]{\includegraphics{figures/letters/#1.pdf}}% panel letter in the figure font, Liberation Sans Bold 8 pt
\usepackage{multirow}%
\usepackage{amsmath,amssymb,amsfonts}%
\usepackage{amsthm}%
\usepackage[title]{appendix}%
\usepackage{xcolor}%
\usepackage{textcomp}%
\usepackage{manyfoot}%
\usepackage{booktabs}%
\usepackage{tabularx}%
\usepackage{xurl}%

\unnumbered%

\usepackage{indentfirst}
\usepackage{etoolbox}

\begin{document}

% Joint supervision marked with a section sign instead of the class default dagger
\makeatletter
\renewcommand{\equalcont}[1]{\gdef\@equalconttext{#1}\g@addto@macro\artauthors{$^{\S}$}\global\equalconttrue}
\patchcmd{\@maketitle}{$^{\dagger}$\@equalconttext}{$^{\S}$\@equalconttext}{}{\PackageWarning{sn-article}{joint-supervision marker patch failed}}
\makeatother

\title[Multilevel report alignment for echocardiography]{EchoAlign: Echocardiography foundation model with multilevel report alignment}

\author*[1]{\fnm{Ahmed} \sur{Aly}}\email{ahmed.aly@mbzuai.ac.ae}

\author[1]{\fnm{Darya} \sur{Taratynova}}

\author[1]{\fnm{Lotfi Abdelkrim} \sur{Mecharbat}}

\author[2]{\fnm{Roman A.} \sur{Sandler}}

\author[2]{\fnm{Joseph Z.} \sur{Sokol}}

\author[2]{\fnm{Joshua} \sur{Penn}}

\author[3]{\fnm{Klaus H.} \sur{Maier-Hein}}

\author[1]{\fnm{Numan} \sur{Saeed}}
\equalcont{These authors jointly supervised this work.}

\author[1]{\fnm{Mohammad} \sur{Yaqub}}
\equalcont{These authors jointly supervised this work.}

\affil*[1]{\orgdiv{Division of Computing and Mathematical Sciences}, \orgname{Mohamed Bin Zayed University of Artificial Intelligence}, \orgaddress{\city{Abu Dhabi}, \country{United Arab Emirates}}}

\affil[2]{\orgname{iCardio.ai}, \orgaddress{\city{Los Angeles}, \postcode{90035}, \state{California}, \country{United States}}}

\affil[3]{\orgdiv{Division of Medical Image Computing}, \orgname{German Cancer Research Center (DKFZ)}, \orgaddress{\city{Heidelberg}, \country{Germany}}}

\abstract{Echocardiography assesses the structure and function of the heart. Cardiologists read each study across several standard views and report the findings visible in those views by cardiac structure. AI foundation models can learn to analyze these studies from clinical reports, reducing annotation burden and supporting clinical interpretation. However, existing models are trained by pairing each video with the full clinical report, including structures and findings its view does not show. This mismatch lets a model associate a finding with incidental or spurious signals rather than its visual evidence, which limits accuracy, anatomical grounding and clinical trust. Here we introduce EchoAlign, a vision-language foundation model developed on 2.87 million echocardiographic videos and images from 159 clinical centers. EchoAlign aligns each cardiac structure, video and study with its report text, and a visibility matrix restricts supervision to the views that actually depict each structure. We compare it with five public foundation models on 0.47 million test videos across ten unseen external datasets from more than 25 clinical centers. EchoAlign achieves the highest mean AUROC of 0.890 (95\% CI, 0.881 to 0.897) for classification and the highest mean quadratic weighted kappa of 0.530 (95\% CI, 0.511 to 0.544) for severity grading. Across 62 clinical findings, EchoAlign provides one model that detects findings, grades their severity, estimates measurements and segments cardiac structures.}

\keywords{}

\maketitle

\begin{figure}[p]%
\centering
\begin{minipage}[t]{72.0mm}
\vspace{0pt}\includegraphics[trim=0.5mm 0 0 0]{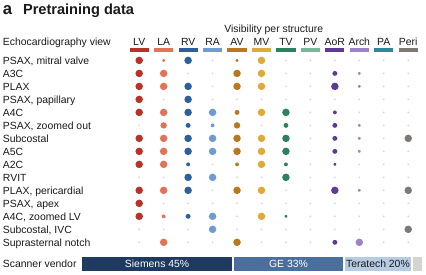}
\end{minipage}\hfill
\begin{minipage}[t]{57.4mm}
\vspace{0pt}\includegraphics[trim=0 0 0.5mm 0]{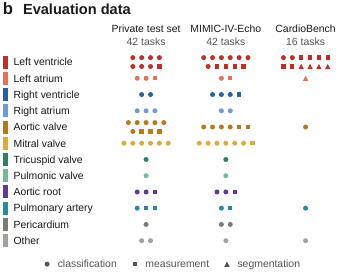}
\end{minipage}\\[1ex]
\begin{minipage}[t]{\textwidth}
\includegraphics{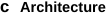}\\[0.6ex]
\includegraphics[width=\linewidth]{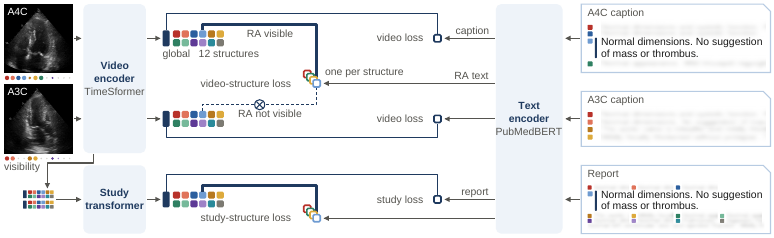}
\end{minipage}\\[1ex]
\begin{minipage}[t]{\textwidth}
\includegraphics{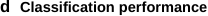}\\[0.6ex]
\includegraphics[width=\linewidth]{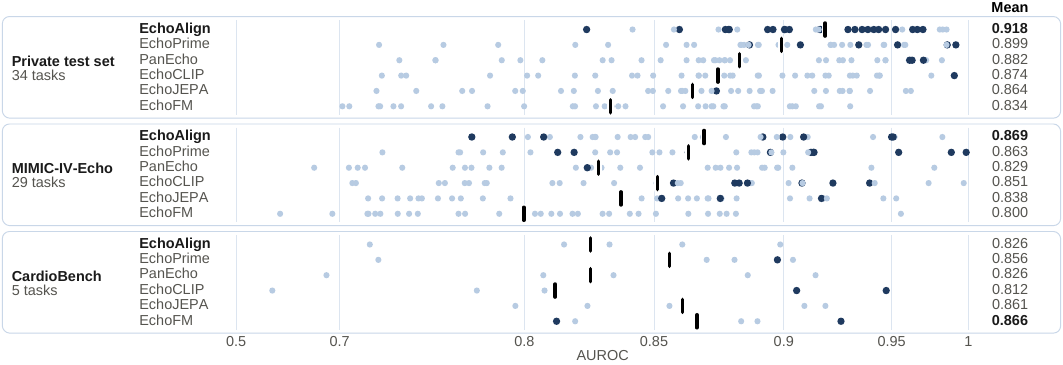}
\end{minipage}
\caption{\textbf{EchoAlign data, tasks, architecture and classification performance.}
\textbf{a}, Pretraining data by cardiac structure, scanner vendor and view. Dot size shows how visible each structure is in that view; Extended Data Fig.~\ref{fig:ed-visibility} gives the full visibility matrix of 12 structures across 66 views.
\textbf{b}, Evaluation tasks by cardiac structure and source; each marker is one task, and its shape shows the task type. Extended Data Fig.~\ref{fig:ed-labels} defines the labels of each task.
\textbf{c}, Architecture and the four contrastive losses of multilevel report alignment; a structure's text trains only the videos whose view shows it.
\textbf{d}, AUROC of the linear probe of each of the six models on each of the 68 classification tasks, by evaluation source, on a logistic axis. Navy dots mark the best model on each task, and black bars mark each model's mean.}\label{fig:overview}
\end{figure}
\clearpage

\section{Introduction}
\label{sec:Introduction}
Cardiovascular diseases accounted for 19.2 million deaths worldwide in 2023 \cite{gbd2023cvd}, and echocardiography is the most widely used imaging modality for assessing cardiac structure and function \cite{douglas2011auc}. A single echocardiographic study shows disease across the chambers, valves and great vessels and allows its severity to be graded, which guides patient management \cite{otto2021guideline}. Because each view shows only part of the heart, a cardiologist reads each study across several standard views and combines them into one clinical report \cite{mitchell2019comprehensive,taub2025guidelines}. This reading is costly in expert time and varies between readers, which motivates automated interpretation by artificial intelligence (AI) \cite{he2023blinded,biner2010reproducibility,thomas2008intraobserver}. Yet most AI models detect one disease at a time, such as reduced ejection fraction \cite{echonetdynamic2019,muhtaseb2022echocotr,maani2024coreecho}, aortic stenosis \cite{wessler2023aortic} or mitral regurgitation \cite{sandler2026mitral}. They learn from expert labels, so every new task inherits the cost of expert reading. Foundation models instead learn a shared embedding from data that clinical care already produces, such as the report written for every study, and adapt it to many diseases and their severities \cite{bommasani2021opportunities,myers2024foundation}.

However, recent benchmarks find that the best echocardiography foundation model changes from one task and dataset to the next \cite{taratynova2025cardiobench}, and no single model yet serves as a general foundation. Existing foundation models fall into three groups. Supervised models such as PanEcho learn a fixed set of tasks from expert labels and map each video directly to those outputs \cite{holste2025complete}. Self-supervised models such as EchoJEPA \cite{munim2026echojepa} and EchoFM \cite{kim2025echofm} learn from unlabeled videos, with objectives that involve neither clinical findings nor text, so each clinical task requires labeled studies. Vision-language models such as EchoCLIP \cite{christensen2024vision} and EchoPrime \cite{vukadinovic2026comprehensive} pair each video with the entire study report and summarize it into one embedding. Expert labels and reports describe the whole study, but each video shows only part of the heart. As a result, a video is trained on findings that it does not show, which dilutes the signal for those it does show. A model can also associate a finding with whatever correlates with it in the video, such as scanner settings, co-occurring diagnoses or patient demographics, rather than with its visual evidence.

Reporting guidelines recommend organizing findings under a heading for each cardiac structure \cite{taub2025guidelines}, so each structure has its own text. Here we introduce EchoAlign, to our knowledge the first echocardiography vision-language model trained to align a separate visual embedding of each cardiac structure with that structure's report and, through a visibility matrix, to learn each embedding only from the views that show its structure (Fig.~\ref{fig:overview}c). These separate embeddings allow a finding to be examined at the level of its cardiac structure. We develop EchoAlign on 2.87 million videos and images from 159 clinical centers (Fig.~\ref{fig:overview}a). We compare it with five public state-of-the-art foundation models on 0.47 million test videos across ten external datasets from more than 25 clinical centers (Fig.~\ref{fig:overview}b). Together, this forms an extensive evaluation of echocardiography foundation models at scale, across the clinical findings that guide patient management. For classification, EchoAlign has the highest mean AUROC with linear probes (Fig.~\ref{fig:overview}d) and, among the models with a text encoder, the highest with text prompts. It also has the highest mean agreement with reported severity grades, estimates cardiac measurements and segments cardiac structures. We release the model and evaluation code publicly to support reproducible research.

\section{Results}
\label{sec:Results}

\textbf{Model and evaluation.} We develop EchoAlign on 2.87 million videos and images from 159 clinical centers. For pretraining, we introduce a novel multilevel report alignment, a contrastive objective \cite{radford2021clip} that matches visual embeddings with report text at three levels: each cardiac structure in every video and study with that structure’s report, each video with a caption built from its report, and each study with its report. We also introduce a visibility matrix that limits each video's supervision to the structures expected to be visible in its view (Fig.~\ref{fig:overview}a,c, Extended Data Fig.~\ref{fig:ed-visibility} and Methods). We compare EchoAlign with the five public foundation models, EchoPrime \cite{vukadinovic2026comprehensive}, PanEcho \cite{holste2025complete}, EchoCLIP \cite{christensen2024vision}, EchoJEPA \cite{munim2026echojepa} and EchoFM \cite{kim2025echofm}, on a private test set from 18 external clinical centers, MIMIC-IV-Echo \cite{PhysioNet-mimic-iv-echo-1.0} and CardioBench \cite{taratynova2025cardiobench}. We evaluate 62 clinical findings: 42 for classification and severity grading, 17 measurements and three segmentation targets (Fig.~\ref{fig:overview}b and Extended Data Fig.~\ref{fig:ed-labels}). Across all datasets, these form four types of task: (1) 68 classification tasks, in a zero-shot setting and with a linear probe, (2) 31 severity grading tasks with an ordinal probe, (3) 25 measurement tasks with a linear probe and (4) segmentation with a trained decoder. Where possible, classification, grading and measurement use one prediction per study, made from all of its videos. Within each task, every model is scored on the same test data and labels (Extended Data Fig.~\ref{fig:ed-data}). The Extended Data tables and figures show the bootstrap 95\% confidence interval of the result for each task \cite{efron1993bootstrap,field2007bootstrapping}. Comparisons between two models use the two-sided $P$ value of a paired bootstrap (Methods).

\begin{figure}[p]%
\centering
\includegraphics[width=\linewidth]{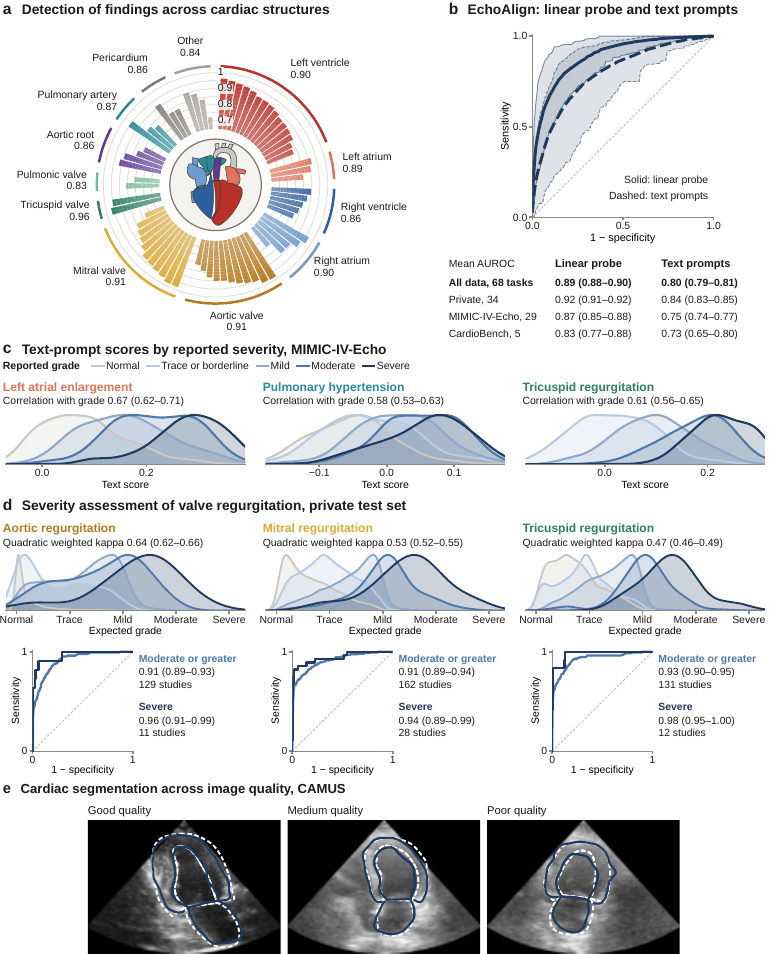}
\caption{\textbf{EchoAlign across cardiac structures, severity grades and segmentation.}
\textbf{a}, AUROC of the linear probe on each of the 68 classification tasks, by cardiac structure.
\textbf{b}, Mean receiver operating characteristic curves of the linear probe and of scores computed from text prompts. Shading shows how much the curves vary between tasks: at each value of 1 $-$ specificity, the middle 80\% of tasks have a sensitivity inside the band.
\textbf{c}, Scores computed from text prompts, by reported grade, on MIMIC-IV-Echo, with Spearman's rank correlation~\cite{spearman1904proof} between score and grade.
\textbf{d}, Expected grade by reported grade with QWK~\cite{cohen1968weighted}, and AUROC for moderate or greater and for severe valve regurgitation, in the private test set.
\textbf{e}, Segmentation contours on CAMUS \cite{leclerc2019deep} images: expert (white, dashed) and EchoAlign (navy).}\label{fig:foundation}
\end{figure}

\textbf{Classification with text prompts.} In the zero-shot setting, each model is used as it is after pretraining, with no further training. For EchoAlign, we use text prompts that describe each finding as present and as absent, and score each finding by the similarity of these prompts to the anatomical study embedding of its own cardiac structure, for example, the mitral valve embedding for mitral regurgitation (Methods). We compare EchoAlign with the three other models that can be scored in this setting. EchoPrime is scored by its report retrieval and EchoCLIP with its released prompts, on the tasks that these methods cover. We extend both models to the other tasks by scoring EchoAlign's or CardioBench's prompts with their own text encoders. PanEcho has no text encoder and is scored with its supervised heads, which are trained on labels for each task, on the tasks that have a head. Across the 68 tasks, EchoAlign's mean AUROC is 0.796, compared with 0.692 for EchoPrime and 0.692 for EchoCLIP ($P < 0.01$ against each; Extended Data Tables~\ref{tab:ed-from-text} and~\ref{tab:ed-cardiobench}a). When EchoPrime and EchoCLIP are each scored only on the tasks that their own method covers, mean AUROC is 0.834 for EchoAlign and 0.716 for EchoPrime on 36 tasks, and 0.831 for EchoAlign and 0.779 for EchoCLIP on 24 tasks ($P < 0.01$ for both). On the 39 tasks that have a PanEcho head, mean AUROC is 0.830 for EchoAlign with text prompts and 0.828 for PanEcho, and the two are not significantly different ($P = 0.579$).

In the private test set, EchoAlign reaches an AUROC of at least 0.900 with text prompts on nine tasks, including left ventricular enlargement and systolic dysfunction, right atrial enlargement, mitral stenosis, and tricuspid and mitral regurgitation. For valve regurgitation, EchoAlign's AUROC is higher than that of EchoPrime and EchoCLIP on all eight tasks ($P < 0.01$), and for tricuspid regurgitation in the private test set it is 0.942, compared with 0.784 for EchoPrime and 0.664 for EchoCLIP. EchoAlign's scores also rise with reported severity (Fig.~\ref{fig:foundation}c). On MIMIC-IV-Echo, for left atrial enlargement, tricuspid regurgitation and pulmonary hypertension, the median score rises with each reported grade, and Spearman's rank correlation~\cite{spearman1904proof} between score and grade is 0.673, 0.609 and 0.580, respectively.

\textbf{Classification with linear probes.} A linear probe is a single linear classifier trained on labels, in the same way for every model, on top of each model's study-level embedding, which for EchoAlign combines its global and 12 anatomical study embeddings (Methods). Across the 68 tasks, EchoAlign's mean AUROC is 0.890, the highest of the six models (Fig.~\ref{fig:overview}d and Fig.~\ref{fig:foundation}b), compared with 0.880 for EchoPrime, 0.860 for EchoCLIP, 0.856 for PanEcho, 0.853 for EchoJEPA and 0.822 for EchoFM ($P < 0.01$ against each; Extended Data Tables~\ref{tab:ed-cardiobench}b and~\ref{tab:ed-classification}). EchoAlign's mean is higher than that of all five models in the private test set ($P < 0.01$) and higher than that of EchoCLIP, PanEcho, EchoJEPA and EchoFM on MIMIC-IV-Echo ($P < 0.01$), and it is not significantly different from EchoPrime on MIMIC-IV-Echo ($P = 0.190$) or from any model on CardioBench ($P \geq 0.116$). EchoAlign ranks first on 32 of the 68 tasks, and its AUROC is at least 0.900 on 32 tasks and at least 0.800 on 65 (Fig.~\ref{fig:foundation}a).

EchoAlign ranks first of the six models on all eight regurgitation detection tasks, with an AUROC of 0.962 for mitral and 0.968 for tricuspid regurgitation in the private test set, and 0.950 and 0.951 on MIMIC-IV-Echo. Its AUROC is higher than that of each of the five models on seven of the eight tasks ($P < 0.01$; Extended Data Table~\ref{tab:ed-classification}), and it is not significantly different from EchoPrime on aortic regurgitation on MIMIC-IV-Echo ($P = 0.054$). In the private test set, EchoAlign is also higher than each of the five models on diastolic dysfunction, heart failure, aortic sclerosis, aortic valve leaflet degeneration, mitral valve thickening, non-trileaflet aortic valve, aortic root dilation, ascending aorta dilation, and left and right atrial enlargement ($P < 0.01$). On 14 tasks, EchoAlign's AUROC is lower than that of the highest model ($P < 0.05$): EchoPrime on aortic stenosis and aortic valve replacement in both test sets, on pacemakers, atrial fibrillation and septal motion abnormality in the private test set, and on mitral valve devices on MIMIC-IV-Echo; EchoCLIP on pericardial effusion, signs of tamponade and right atrial enlargement on MIMIC-IV-Echo and on pulmonary arterial hypertension on CardioBench; and EchoJEPA on pulmonary hypertension and right ventricular systolic dysfunction on MIMIC-IV-Echo, on which it is pretrained.

Across age groups in the private test set, over 11 findings, EchoAlign's mean AUROC exceeds that of EchoPrime, the next-best model, by 0.016 at 40 to 59 years ($P < 0.05$) and by 0.024 at 60 to 79 years ($P < 0.01$), and the two models are not significantly different below 40 years ($P = 0.060$) or at 80 years or older ($P = 0.646$). On MIMIC-IV-Echo, over 21 findings, EchoAlign exceeds EchoPrime by 0.010 at good or adequate image quality ($P < 0.05$), and the two are not significantly different at suboptimal or poor image quality ($P = 0.738$; Extended Data Fig.~\ref{fig:ed-subgroups}).

\textbf{Severity grading.} We grade each finding on its ordered severity scale (Extended Data Fig.~\ref{fig:ed-labels}a) with an ordinal probe, a multinomial logistic regression trained on the same study-level embedding as the linear probes. We measure agreement with the reported grade by quadratic weighted kappa (QWK)~\cite{cohen1968weighted}. Across the 31 graded tasks, EchoAlign's mean QWK is 0.530, the highest of the six models, compared with 0.505 for EchoPrime, 0.477 for EchoCLIP, 0.457 for PanEcho, 0.441 for EchoJEPA and 0.389 for EchoFM ($P < 0.01$ against each; Extended Data Table~\ref{tab:ed-regression-grading}b). In the private test set, EchoAlign's mean QWK is 0.473, higher than that of each of the five models ($P < 0.01$). On MIMIC-IV-Echo, it is 0.562, higher than that of EchoCLIP, PanEcho, EchoJEPA and EchoFM ($P < 0.01$) and not significantly different from 0.561 for EchoPrime ($P = 0.930$).

EchoAlign ranks first of the six models on all six regurgitation grading tasks, with a QWK of 0.643 for aortic, 0.534 for mitral and 0.472 for tricuspid regurgitation in the private test set, and 0.619, 0.670 and 0.710 on MIMIC-IV-Echo. Its QWK is higher than that of each of the five models on each of these tasks ($P < 0.01$). In the private test set, EchoAlign is also higher than each of the five models on left atrial enlargement, diastolic dysfunction, left ventricular hypertrophy, mitral annular calcification and pulmonary hypertension ($P < 0.05$). On seven MIMIC-IV-Echo tasks, EchoAlign's QWK is lower than that of the highest model ($P < 0.05$): EchoPrime and EchoCLIP on three each and EchoJEPA on one (Extended Data Table~\ref{tab:ed-regression-grading}b). The probe also gives each study an expected grade (Fig.~\ref{fig:foundation}d and Methods). In the private test set, the median expected grade rises with each reported grade for aortic, mitral and tricuspid regurgitation, and the expected grade separates severe regurgitation from lower grades with an AUROC of 0.962, 0.944 and 0.977 (11, 28 and 12 severe studies).

\textbf{Measurements and segmentation.} We estimate measurements with ridge regression probes on the study-level embedding. Across the 25 measurement tasks, EchoPrime has the lowest root-mean-square error (RMSE) on 11 tasks and EchoAlign on six (Extended Data Tables~\ref{tab:ed-cardiobench}c and~\ref{tab:ed-regression-grading}a). For ejection fraction, EchoAlign's RMSE is 5.02 percentage points in the private test set, 7.92 on MIMIC-IV-Echo and 7.40 on EchoNet-Dynamic \cite{echonetdynamic2019}, compared with 4.77 and 6.73 for EchoPrime and 6.49 for EchoJEPA, the lowest in each dataset ($P < 0.01$). For aortic root diameter in the private test set, EchoAlign's RMSE is lower than that of each of the five models ($P < 0.05$). For segmentation, we train a convolutional decoder on each model's fixed features. EchoAlign reaches a mean Dice similarity coefficient~\cite{dice1945measures,sorensen1948method} of 0.880 on CAMUS \cite{leclerc2019deep} across the left ventricular cavity, left ventricular myocardium and left atrium, second to 0.892 for EchoFM (Fig.~\ref{fig:foundation}e and Extended Data Table~\ref{tab:ed-cardiobench}d). For the left ventricular cavity, it reaches 0.917 on EchoNet-Dynamic and 0.896 on EchoNet-Pediatric \cite{echonetpediatric2025}, and 0.813 and 0.767 with decoders trained on CAMUS (Extended Data Table~\ref{tab:ed-cardiobench}e).

\textbf{Data scaling.} EchoPrime, the model with the next-highest mean AUROC and QWK, is pretrained on about 4.5 times as many studies as EchoAlign, 272,256 compared with 60,782 \cite{vukadinovic2026comprehensive}. To compare the two pretraining recipes on equal data, we pretrain EchoAlign and a reproduction of the EchoPrime recipe on the same subsets of studies (Extended Data Fig.~\ref{fig:ed-efficiency}a). The reproduction uses EchoPrime's published encoders, initialization and optimizer settings, and is scored with EchoPrime's released view classifier and section weights (Methods). Across the 68 tasks, as the subsets grow from 3,000 to 10,000 and 30,000 studies, mean AUROC rises from 0.815 to 0.839 and 0.869 for EchoAlign and from 0.725 to 0.745 and 0.803 for the EchoPrime recipe. With 1\% of the training labels for the probes, the released models reach a mean AUROC of 0.888 for EchoAlign and 0.870 for EchoPrime across 32 tasks in the private test set, and 0.826 and 0.827 across 22 tasks on MIMIC-IV-Echo (Extended Data Fig.~\ref{fig:ed-efficiency}b).

\section{Discussion}
\label{sec:Discussion}

EchoAlign is an echocardiography vision-language model that learns each cardiac structure from the part of the report that describes it and only from the views that show it. Evaluated against five public foundation models on ten external datasets, it achieves the highest mean AUROC for classification and the highest mean QWK for severity grading, while also estimating cardiac measurements and segmenting cardiac structures. It does so with fewer pretraining studies than EchoPrime, the next-best model, and its mean AUROC remains higher when both pretraining recipes are given the same studies (Extended Data Fig.~\ref{fig:ed-efficiency}a). 

Previous echocardiography vision-language models pair each video with the entire study report and summarize it into one embedding \cite{christensen2024vision,vukadinovic2026comprehensive}, whereas EchoAlign provides a separate anatomical study embedding for each cardiac structure, which relates its predictions to cardiac anatomy. A finding is read better from the embedding of its own cardiac structure than from the average of the 12 embeddings (Extended Data Fig.~\ref{fig:ed-token-map}), indicating that each embedding encodes the findings of the structure to which it was aligned. A score obtained from text prompts is thus derived from a single, named cardiac structure, whereas the linear probes, which combine all embeddings, benefit from the complementary information that the other structures provide.

Echocardiographic reading varies between readers \cite{he2023blinded,biner2010reproducibility,thomas2008intraobserver}, and because the severity grade guides patient management \cite{otto2021guideline}, a grade that shifts with the reader can alter the care that a patient receives. EchoAlign applies one rule to every study and grades each finding on its ordered severity scale, offering a consistent second read from the same model that detects the other findings of the study. Its detection and grading of valve regurgitation hold in both the private test set and MIMIC-IV-Echo, which were acquired at different clinical centers, and the grading extends to other findings reported on an ordered scale, among them left atrial enlargement, diastolic dysfunction and pulmonary hypertension. Because cardiologists already write a report for every study, supervision requires no additional annotation, and a finding can be scored directly from text prompts that describe it, without task-specific training.

This study is retrospective, and a prospective trial, as for EchoPrime \cite{kumar_2026_20359691}, is the next step to measure EchoAlign's effect on missed diagnoses and reading time. Other limitations include the wide confidence intervals on CardioBench, as HMC-QU \cite{degerli2021early,degerli2023multiview} comprises 50 videos from 25 patients and SegRWMA \cite{liu2023enhance} four positive patients among 56. Scores obtained from text prompts depend on the wording of those prompts, as in any zero-shot setting \cite{radford2021clip}. In the private test set, spectral Doppler and M-mode recordings are not provided as inputs, so that targets such as transaortic gradients and peak velocity are estimated from B-mode and color Doppler videos alone.

EchoAlign provides one model for classifying echocardiographic findings with or without task-specific training, grading their severity, estimating cardiac measurements and segmenting cardiac structures. It learns from the reports that cardiologists already write in routine care, matching the findings of each cardiac structure to the views that show it. Routine clinical reports, through their structure as much as their number, can therefore supervise a single foundation model that spans the findings guiding patient management in echocardiography.

\clearpage
\bibliography{sn-bibliography}%

\clearpage
\section{Methods}
\label{sec:Methods}

\subsection{Study design}\label{sec:study-design}

We pretrain EchoAlign on clinical echocardiography studies and their reports from 150 clinical centers (Fig.~\ref{fig:overview}a). We evaluate it on a private test set from 18 external clinical centers, MIMIC-IV-Echo and CardioBench (Fig.~\ref{fig:overview}b). We compare EchoAlign with five publicly available echocardiography foundation models on 68 classification tasks, 31 graded tasks, 25 measurement tasks and three segmentation targets, which cover 62 clinical findings. Every model is scored on the same findings across three evaluation sources.

\subsection{Model architecture and objectives}\label{sec:model-training}

\begin{figure}[!h]
\centering
\includegraphics[width=\linewidth]{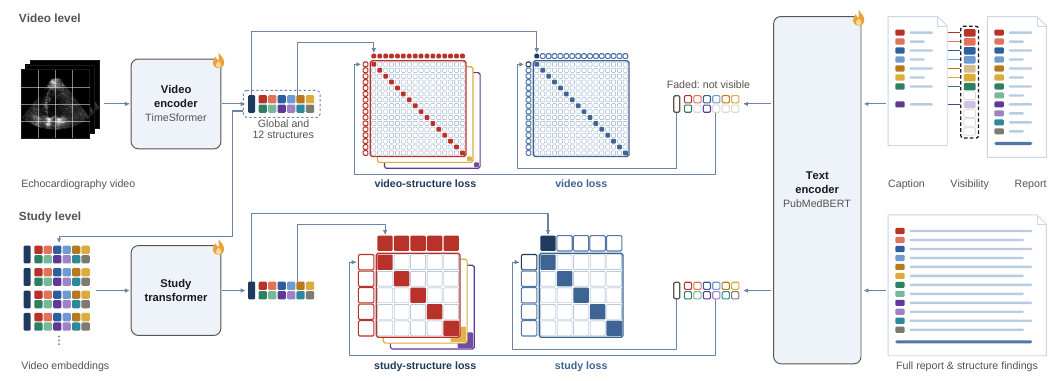}
\caption{\textbf{Architecture and multilevel report alignment.} Each video has a global embedding and 12 anatomical embeddings, and the study transformer combines the videos of a study into study embeddings. Four contrastive losses match each embedding with its text; a faded structure is not visible in the view and has no text.}\label{fig:architecture}
\end{figure}

EchoAlign learns from videos and reports by CLIP-style contrastive learning \cite{radford2021clip}. It pulls the embeddings of matched visual and text inputs together in a shared 512-dimensional space and pushes mismatched ones apart. We apply this contrastive learning at multiple levels: the video, the cardiac structure and the study (Fig.~\ref{fig:architecture}). Each video has a global embedding and 12 anatomical embeddings, one per cardiac structure. Each study has a global embedding and 12 anatomical embeddings, which combine all its videos. We first describe the text that the embeddings are matched with, then the encoders and the study transformer that produce the embeddings, and finally the four losses that match them.

\begin{figure}[t]
\centering
\includegraphics[width=\linewidth]{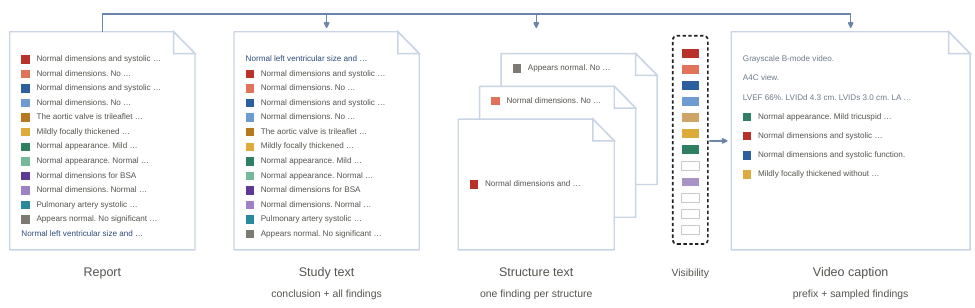}
\caption{\textbf{Text built from one report.} The report of one training study and the texts built from it, with truncated text.}\label{fig:text-from-reports}
\end{figure}

\textbf{Text from reports.} Each study has a clinical report with a conclusion and findings for 12 cardiac structures (Fig.~\ref{fig:text-from-reports}, Report). From it, we build three kinds of text (Fig.~\ref{fig:architecture}). Study text concatenates the conclusion and all non-empty findings in a fixed order, and we match it with the study (Fig.~\ref{fig:text-from-reports}, Study text). Structure text is the finding for one structure, and we match it with the anatomical embedding of that structure (Fig.~\ref{fig:text-from-reports}, Structure text). A caption is the text that we match with a video (Fig.~\ref{fig:text-from-reports}, Video caption). The text encoder accepts at most 256 tokens; for longer text, we sample a contiguous window of up to five sentences.

A video shows only some cardiac structures, so its caption should describe only the structures that its view shows. We introduce a view-by-structure visibility matrix of 66 views (Extended Data Fig.~\ref{fig:ed-visibility}) that gives each view-structure pair a training weight $w_{vk}\in\{0,0.1,0.3,0.5,0.8,1.0\}$, the expected visibility of structure $k$ in view $v$. Fig.~\ref{fig:text-from-reports} (Visibility) shows the weights of one view: a full color is a weight of 1, a lighter color is a smaller weight and an empty square is a weight of 0. We set the weights using American Society of Echocardiography guidelines as the primary reference \cite{mitchell2019comprehensive}. Otherwise, we use secondary references \cite{ahmed2026echoimaging,lankford2022psax} and representative examples from our dataset. The matrix encodes expected visibility; the report states what was observed. The dataset provides the view labels, and we use them only in pretraining.

A caption begins with short sentences stating the modality, the view when known and the available study measurements: ejection fraction, left ventricular end-diastolic and end-systolic diameters, and left atrial dimension (Fig.~\ref{fig:text-from-reports}, Video caption). Findings follow. For each batch, we draw a threshold uniformly from zero to one, and the caption contains the non-empty findings whose visibility weight exceeds it, so a more visible structure enters captions more often. When a caption has more than one finding, we then drop each finding with probability 0.2, keep at least one and shuffle the order of the rest. These random choices give a video a different caption each time we sample it. In the example of Fig.~\ref{fig:text-from-reports}, the draw selects the six findings with weight 1, dropping removes the left and right atria, and shuffling sets the order of the remaining four. A caption is the report conclusion instead when no finding exceeds the threshold, or when another video in the batch has the same report text and the report has a conclusion.

\textbf{Encoders and embeddings.} The video encoder is a TimeSformer, a video transformer that extends the Vision Transformer (ViT) architecture \cite{dosovitskiy2021vit} with divided space-time attention, initialized from Kinetics-400-pretrained \cite{kay2017kineticshumanactionvideo} TimeSformer weights \cite{bertasius2021timesformer}. The text encoder is a BERT model \cite{devlin2019bert}, initialized from PubMedBERT weights pretrained on PubMed abstracts and full text \cite{gu2021pubmedbert}. For each encoder, a bias-free linear projection maps the 768-dimensional class-token output to 512 dimensions, followed by $\ell_2$ normalization. For a video, the result is its global embedding. The same text encoder and projection give the embeddings of captions, study texts and structure texts. We fine-tune both encoders and their projections during pretraining.

To obtain 12 anatomical embeddings for each video, we add 12 learned anatomical queries and an eight-head cross-attention layer on top of the TimeSformer. For video $i$, let $\mathbf{H}_i\in\mathbb{R}^{N\times768}$ denote the matrix of its $N=3{,}136$ space-time patch tokens (196 patches in each of 16 frames, without the class token) and $\mathbf{Q}\in\mathbb{R}^{12\times768}$ the learned queries. Every query attends to all patch tokens of the video, without a mask. The cross-attention layer, multi-head attention (MHA) \cite{NIPS2017_3f5ee243} with its learned projections, gives the anatomical embeddings
\begin{equation*}
\mathbf{A}_i=\operatorname{MHA}(\mathbf{Q},\mathbf{H}_i,\mathbf{H}_i),\qquad
\mathbf{z}_{ik}=\frac{\mathbf{P}_{q}\operatorname{LN}(\mathbf{a}_{ik})}{\lVert\mathbf{P}_{q}\operatorname{LN}(\mathbf{a}_{ik})\rVert_2},
\end{equation*}
where $\mathbf{a}_{ik}^{\top}$ is row $k$ of $\mathbf{A}_i$, $\operatorname{LN}$ denotes layer normalization and $\mathbf{P}_{q}\in\mathbb{R}^{512\times768}$ is a bias-free projection. Nothing in this architecture ties query $k$ to a structure. The video-structure loss defined below trains query $k$ to match the structure text of structure $k$.

A report describes a whole study, but the video encoder sees one video at a time. We therefore add a study transformer on top of the video embeddings, which combines the videos of a study into one global study embedding and 12 anatomical study embeddings. The study and study-structure losses defined below match these embeddings with the study text and the structure texts. The input tokens of the study transformer are the global embedding and the 12 anatomical embeddings of each video, each with 512 dimensions. During training, it receives the one to eight videos sampled from the study for the batch. Additive embeddings mark the video type (B-mode video, color Doppler video, still image or unknown) and the role of each token (global or one of the 12 structures). Two self-attention layers \cite{NIPS2017_3f5ee243}, each with four heads, a 1,024-unit feed-forward layer and dropout 0.1, exchange information across the tokens. This self-attention has no mask other than padding, so every token attends to every other token of the study. Videos carry no positional encoding, so the study transformer treats a study as an unordered set of videos. Thirteen learned queries in a four-head cross-attention layer then pool the tokens. One query pools the global tokens into the global study embedding, and each of the other 12 queries pools the tokens of its structure into the anatomical study embedding of that structure. Each study embedding has 512 dimensions.

The pooling queries use the visibility matrix as well. During training, each anatomical query attends to the tokens of its structure from videos whose view has positive visibility for that structure. Videos with no known view are masked out of these queries but still enter self-attention and the global query. Because the study transformer receives no view information at inference, each video has a 10\% chance during training of being treated in this pooling as having every structure visible.

\textbf{Multilevel report alignment.} Four contrastive losses match each kind of embedding with its text (Fig.~\ref{fig:architecture}). Each is the standard CLIP loss \cite{radford2021clip} on normalized embeddings, with one learned logit scale, shared by all four losses and constrained to $[1,32]$.

The video loss $\mathcal{L}_{\mathrm{video}}$ matches each global video embedding with the embedding of its caption. Pairs with identical captions are removed from the negatives; videos from the same study otherwise remain negatives. The video-structure loss $\mathcal{L}_{\mathrm{video\text{-}structure}}$ matches each eligible anatomical embedding with the structure text of its structure, excluding identical findings as negatives. An anatomical embedding is eligible when its view has $w_{vk}>0$ and the report contains a finding for structure $k$, so a video with no known view is masked out of this loss. We average this loss over structures with at least two eligible videos in the batch. The study loss $\mathcal{L}_{\mathrm{study}}$ matches each global study embedding with its study text, and studies with identical text form one positive set with equal target probability. The study-structure loss $\mathcal{L}_{\mathrm{study\text{-}structure}}$ applies the structure formulation to anatomical study embeddings. It includes a structure when at least one video of the study in the batch has positive visibility for it, so a structure without a report finding is matched with the encoding of an empty finding. An anatomical study embedding with no such video is left out of this loss. The alignment objective is
\begin{equation*}
\mathcal{L}_{\mathrm{align}}=\mathcal{L}_{\mathrm{video}}+
0.5\mathcal{L}_{\mathrm{video\text{-}structure}}+
0.5\mathcal{L}_{\mathrm{study}}+
0.1\mathcal{L}_{\mathrm{study\text{-}structure}}.
\end{equation*}

\subsection{Pretraining}\label{sec:pretraining}

We pretrain EchoAlign on 60,782 echocardiographic studies across 150 clinical centers served by telecardiology companies (Fig.~\ref{fig:overview}a), after excluding studies with corrupted reports. The pretraining data comprise 2,651,674 videos and still images: 1,033,450 B-mode videos, 832,319 color Doppler videos and 785,905 still images. The still images include spectral Doppler and M-mode recordings. The validation set used during pretraining adds 217,195 videos from 9 other clinical centers, for 2,868,869 videos and images from 159 clinical centers.

Each study combines videos from multiple views with a clinical report, which supplies a conclusion and findings for 12 cardiac structures: the left and right ventricles, the left and right atria, the aortic, mitral, tricuspid and pulmonic valves, the pericardium, the aortic root, the aortic arch and the pulmonary artery (Fig.~\ref{fig:overview}a). We use these existing report fields as training text, without additional annotation.

We resample each video to 0.36 mm per pixel, remove the annotation text box, extract the ultrasound sector, normalize intensities, pad to a square and resize to $224\times224$ pixels with three channels (Extended Data Fig.~\ref{fig:ed-data}b). We also resample videos to one common frame rate by nearest-frame selection, because acquisition frame rates vary. Still images enter the first stage as single-frame inputs to the same encoder, so temporal attention sees one frame and no repeated frames. For the anatomical queries, we repeat the 196 patch tokens of a still across the 16 frame positions, which leaves their attention unchanged.

We sample 16 consecutive frames per video, from a random start in training and from the centered window in validation. Videos shorter than 16 frames repeat their last frame. We interpolate the TimeSformer's eight temporal-position embeddings to 16 frames, so temporal attention spans all 16 frames and spatial attention stays within each frame. Training applies one geometric transform to all frames of a video: rotation up to 20 degrees, translation up to 10\%, scaling between 0.85 and 1.15 and perspective displacement up to 12\% of frame size.

In each epoch, we shuffle each study's videos and fill each batch with chunks of one to eight videos, drawn at random from the studies, so the study transformer integrates up to eight videos of a study in a batch. A study can return in later batches of the epoch. Training runs on 64 AMD Instinct MI210 GPUs with a batch size of 16 per GPU (global batch 1,024) and gradient clipping at 1.0. Each study is assigned to one GPU, and each GPU builds its own batches and draws its own caption threshold. The losses gather embeddings across GPUs, and every GPU runs the same number of batches per epoch, by dropping extra videos or repeating complete batches.

The first stage trains on videos and still images for 10 epochs with the Lion optimizer \cite{chen2023lion}, a learning rate of $8\times10^{-5}$, weight decay 0.1, one warm-up epoch and cosine decay to zero. We keep the first-stage checkpoint at the end of the third epoch, which had the lowest validation loss $\mathcal{L}_{\mathrm{align}}$. The second stage continues from this checkpoint for 10 further epochs on videos of at least 16 frames, with a learning rate of $8\times10^{-6}$, no weight decay and a warm-up of 0.05 epoch. EchoAlign is the final second-stage checkpoint.

\subsection{Datasets and baseline models}\label{sec:cohorts}

The private validation set comes from 9 other clinical centers and the private test set from 18 external clinical centers. They contain 6,811 and 11,797 studies, respectively, with 362,374 videos in the test set (Extended Data Fig.~\ref{fig:ed-data}a). The private training set contains 60,782 studies. Labels come from structured report fields and from pattern matching on reports.

We define 34 classification tasks covering cardiac morphology and function, hemodynamic findings, clinical composites, device and intervention markers, and technical study assessment (Extended Data Fig.~\ref{fig:ed-labels}). Each task has a fixed binary boundary, chosen for clinical relevance, that separates studies at or above a grade, such as aortic regurgitation of at least moderate severity, from all other studies. Eight measurement targets cover ejection fraction, left atrial volume index, pulmonary artery pressure, right ventricular systolic pressure, aortic root diameter, and transaortic mean gradient, peak gradient and peak velocity (Extended Data Fig.~\ref{fig:ed-labels}b). Eleven findings are also graded on their ordered severity scale (Extended Data Fig.~\ref{fig:ed-labels}a). Pulmonary hypertension refers to the echocardiographic estimate recorded in the report.

\textbf{MIMIC-IV-Echo.} MIMIC-IV-Echo v1.0 contains de-identified transthoracic studies acquired at Beth Israel Deaconess Medical Center from 2017 to 2019 on GE Vivid E90, E95 and S70 systems \cite{PhysioNet-mimic-iv-echo-1.0}. We analyze 525,319 videos from 7,243 studies in 4,579 patients. Labels come from the clinician-verified structured measurement record linked to each study. We binarize graded fields at the boundary of the corresponding task in the private test set. MIMIC-IV-Echo gives 29 classification, 20 graded and 11 measurement tasks (Extended Data Fig.~\ref{fig:ed-labels}). Data are split by patient into 5,078 training studies from 3,224 patients, 662 validation studies from 443 patients and 1,503 test studies from 912 patients (109,111 videos).

\textbf{CardioBench.} We evaluate 11 CardioBench dataset-task pairs from eight datasets, using the released splits (Extended Data Figs.~\ref{fig:ed-labels} and~\ref{fig:ed-data}a) \cite{taratynova2025cardiobench}. TMED-2 contributes labeled images \cite{huang2022tmed}, CAMUS two- and four-chamber videos \cite{leclerc2019deep}, SegRWMA apical two-, three- and four-chamber (A2C, A3C and A4C) videos \cite{liu2023enhance}, HMC-QU A2C and A4C videos \cite{degerli2021early,degerli2023multiview} and EchoNet-Pediatric A4C and parasternal short-axis videos \cite{echonetpediatric2025}. EchoNet-Dynamic and EchoNet-LVH provide one video per study \cite{echonetdynamic2019,echonetlvh2020}, and CardiacNet one A4C video per source case \cite{yang2024cardiacnet}. The data come from Tufts Medical Center (TMED-2), the University Hospital of St Etienne (CAMUS), Hamad Medical Corporation (HMC-QU), Lucile Packard Children's Hospital (EchoNet-Pediatric), Stanford Medicine (EchoNet-Dynamic and EchoNet-LVH) and four hospitals (CardiacNet). We make one prediction per study for TMED-2 and EchoNet-Pediatric and per patient for SegRWMA and CAMUS. For the other seven pairs, we treat each released video as one study. We exclude 14 SegRWMA files whose image volumes hold segmentation masks rather than echocardiograms. The CardioBench test sets contain 2,734 videos and 3,602 TMED-2 images. With the private test set and MIMIC-IV-Echo, the ten test datasets contain 474,219 videos, and none of the MIMIC-IV-Echo or CardioBench data are used in pretraining.

\textbf{Baseline models.} We compare EchoAlign with five publicly available echocardiography foundation models. EchoPrime uses a Multiscale Vision Transformer \cite{fan2021multiscalevisiontransformers} video encoder with a 512-dimensional output, pretrained on 272,256 studies and 11,984,170 videos from Cedars-Sinai, and we load the released encoder with its view classifier \cite{vukadinovic2026comprehensive}. PanEcho uses a ConvNeXt-Tiny \cite{liu2022convnet2020s} frame encoder and a four-layer, eight-head temporal transformer, developed on 32,265 studies and 1.2 million videos from Yale New Haven, and we take the 768-dimensional features before its supervised task heads \cite{holste2025complete}. EchoCLIP uses a ConvNeXt-Base image encoder with a 512-dimensional output, developed on 1,032,975 video-report pairs from 224,685 Cedars-Sinai studies, 195,082 of them in its training set, and loaded from the released weights \cite{christensen2024vision}. EchoJEPA uses a ViT-L video encoder pretrained with latent prediction. Its authors report a larger ViT-G model trained on 18 million echocardiograms from 300,000 patients, but have not released it, and every released EchoJEPA model is pretrained on MIMIC-IV-Echo, about 525,000 videos \cite{munim2026echojepa}. We use the ViT-L model its authors recommend for functional tasks. Its pretraining therefore included, without labels, the MIMIC-IV-Echo videos in our test set, so its MIMIC-IV-Echo results are not an external evaluation. EchoFM uses a ViT-L video encoder pretrained with masked autoencoding on two public datasets and an internal dataset of 286,080 videos from 6,500 patients \cite{kim2025echofm}. EchoAlign is pretrained on 60,782 studies and 2,651,674 videos and images. EchoJEPA and EchoFM have no text encoder, so we do not score them with text prompts.

All models receive frames in standard display orientation, and each model keeps its own input preprocessing, documented in the public repository. On the private dataset, all models read the same stored frames. EchoAlign reads 16 consecutive frames from the start of each video at its training frame rate and repeats a single image into a 16-frame clip. At the stored frame rate, EchoPrime reads every second frame of the first 32, PanEcho 16 consecutive frames, EchoCLIP every second frame of the first 40 and EchoFM 32 consecutive frames. EchoJEPA reads 16 consecutive frames at 8 frames per second. As in its released code, EchoPrime embeds videos only; for TMED-2, whose files are images, it repeats each image into a clip. EchoCLIP encodes each frame, and we average and renormalize the frame embeddings. PanEcho, EchoCLIP, EchoJEPA and EchoFM average video embeddings within each prediction unit. EchoPrime uses its own study pooling: the view-weighted embedding of the report section that matches each task, or the mean of its 15 section embeddings for tasks without a matching section.

\subsection{Evaluation}\label{sec:eval-tasks}

Linear probes read fixed features, so we first turn each prediction unit into one feature vector, its study-level embedding. A prediction unit is a study or, for some CardioBench datasets, a patient or a single video. For EchoAlign, we compute the vector in four steps, and every fitted probe uses it.

First, the video encoder gives each video of the unit a global embedding and 12 anatomical embeddings. Second, the pretrained study transformer combines these into 13 study embeddings of 512 dimensions each, one global and 12 anatomical. At inference, in all three evaluation sources, the study transformer receives no view information. Every video has the unknown video type, and each anatomical query attends to its structure's tokens from all videos. Evaluation therefore needs no view labels.

Third, for each of the 13 study embeddings, we fit a principal-component projection with 64 components on the private training studies, without labels. Fourth, we concatenate the 13 projections, which gives 13 $\times$ 64 = 832 features. We fit the projections once and reuse them unchanged for MIMIC-IV-Echo and CardioBench. In the data scaling analysis, each subset model has its own study transformer, and we refit its projections on the private training set.

\textbf{Classification and measurement probes.} Linear probes measure the clinical information that each model's study-level embedding contains. We evaluate 62 clinical findings: 42 for classification and severity grading, 17 measurements and three segmentation targets (Fig.~\ref{fig:overview}b). Of the 42, 41 have a classification task, and ejection fraction category is graded only. A finding evaluated in several datasets counts once, and each dataset keeps its own label definition.

For each task, we standardize features, fit candidate probes on the training set and select regularization on the validation set. We then refit standardization and the selected probe on training and validation data together. Classification uses L2-regularized logistic regression, with $C\in\{10^{-4},10^{-3},0.01,0.1,1,10,100\}$ selected by validation area under the receiver operating characteristic curve (AUROC). Measurement uses ridge regression, with $\alpha\in\{0.01,0.1,1,10,100,10^{3},10^{4}\}$ selected by validation root-mean-square error (RMSE).

\textbf{Severity grading.} We evaluate severity grading in the private test set and on MIMIC-IV-Echo for findings graded on an ordered scale, and each finding keeps its native scale (Extended Data Fig.~\ref{fig:ed-labels}a and Extended Data Table~\ref{tab:ed-regression-grading}b). We exclude ungraded positive labels, compound grades such as mild to moderate, and studies marked as not assessable. MIMIC-IV-Echo records aortic, mitral and tricuspid regurgitation, pericardial effusion and mitral annular calcification only when present, so we grade them from their lowest recorded grade. When a grade of a MIMIC-IV-Echo finding is missing from a set, its top two grades merge; we omit a finding that still lacks a grade in a set, or whose merged grade has fewer than five test studies.

The ordinal probe is a standardized multinomial logistic regression on the same features as the classification probes. We select $C$ from the classification grid by validation quadratic weighted kappa (QWK) and refit on training and validation data. The predicted grade probabilities give an expected-grade score, and cut points that maximize QWK on training and validation data convert this score to a grade. We also report the AUROC of the expected-grade score for separating moderate or greater regurgitation, the binary boundary, and severe regurgitation from lower grades. QWK penalizes each disagreement with the reported grade by the square of its distance in grades.

\textbf{Classification with text prompts.} In the zero-shot setting, we score each study by its similarity to text prompts that describe a finding as present and as absent. For a normalized 512-dimensional visual embedding $\mathbf{x}$ and the sets of present and absent prompts $\mathcal{P}_{+}$ and $\mathcal{P}_{-}$, encoded by EchoAlign's pretrained text encoder, the score is
\begin{equation*}
s(\mathbf{x})=\frac{1}{\lvert\mathcal{P}_{+}\rvert}\sum_{\mathbf{t}\in\mathcal{P}_{+}}\operatorname{cos}(\mathbf{x},\mathbf{t})-\frac{1}{\lvert\mathcal{P}_{-}\rvert}\sum_{\mathbf{t}\in\mathcal{P}_{-}}\operatorname{cos}(\mathbf{x},\mathbf{t}).
\end{equation*}
The score uses the anatomical study embedding of the cardiac structure assigned to the finding, or the global study embedding when the finding has no structure, before principal-component projection.

Only the prompts are selected on the private validation set: for each finding, we select up to three short sentences per class from curated candidates. We transfer these prompts unchanged to the MIMIC-IV-Echo and CardioBench tasks that repeat a private finding. CardioBench tasks without a private match use CardioBench's own prompts. The public repository contains all prompts.

The text-based baselines are scored by their own methods wherever these cover a task: EchoCLIP uses its released prompts and EchoPrime its report retrieval, as in its released code, which averages the labels of the 50 most similar reports. For the other tasks, both use EchoAlign's or CardioBench's prompts, encoded by their own text encoders. PanEcho is scored with its supervised heads on the 39 tasks they cover.

\textbf{Anatomical study embedding analysis.} The anatomical study embedding of a finding's assigned structure contains information about that finding. On the private validation set, we rank the 13 study embeddings for each finding by AUROC with text prompts. The assigned embedding, which is the global study embedding for the two findings without an assigned structure, ranks first on 14 of 34 findings and in the top three on 26, with a median rank of 2. For Extended Data Fig.~\ref{fig:ed-token-map}, we fit a linear probe on each 512-dimensional study embedding separately, for all classification tasks of the private test set (34) and MIMIC-IV-Echo (29), on the test studies of the performance tables. We express each AUROC relative to the mean AUROC of the 12 anatomical study embeddings for that finding. For the 32 and 25 findings with an assigned structure, we compare the anatomical study embedding of the finding's own structure with that mean and with the 832-dimensional study-level embedding, using paired bootstrap differences.

\textbf{Data scaling.} We compare the EchoAlign recipe with an EchoPrime reproduction on nested, center-stratified subsets of 3,000, 10,000 and 30,000 pretraining studies, with the same studies for both recipes at each budget (Extended Data Fig.~\ref{fig:ed-efficiency}a). The reproduction initializes its video encoder from Kinetics-400-pretrained \cite{kay2017kineticshumanactionvideo} MViT-v2-S weights \cite{li2022mvitv2improvedmultiscalevision} and its text encoder from PubMedBERT weights pretrained on PubMed abstracts only \cite{gu2021pubmedbert}, and receives no weights from the released EchoPrime model for training; its probes use the released view classifier and section weights. It uses the published optimizer settings (AdamW \cite{loshchilov2019decoupledweightdecayregularization}, learning rate $4\times10^{-5}$, weight decay $10^{-6}$, first six text-encoder layers frozen, 60 epochs). It departs from the published recipe in two ways: the logit scale starts at $1/0.07$, and the fine-tuning on cleaned reports is omitted. EchoAlign uses its TimeSformer and PubMedBERT initialization, still images and multilevel report alignment, with a batch of 1,024. Following the published training description, in which each batch holds 32 video-report pairs from different studies \cite{vukadinovic2026comprehensive}, the reproduction samples one video per study per epoch, with a batch of 32. The recipes therefore share unique study budgets but differ in video exposure, optimization and compute.

For EchoAlign, we use the final checkpoint of the second stage and refit its principal-component projection on each run. For the reproduction, we select the checkpoint with the lowest pretraining validation loss and use section pooling. Logistic-regression probes on the pretrained features select regularization on the validation set. We report 34 private-test-set, 29 MIMIC-IV-Echo and five CardioBench classification tasks, averaged with equal weights. Confidence intervals use the bootstrap described in the Statistical analysis section.

\textbf{Label efficiency.} We repeat the classification probes of the six models with 1\%, 5\%, 10\%, 25\%, 50\% and 100\% of the training labels, in the private test set and on MIMIC-IV-Echo (Extended Data Fig.~\ref{fig:ed-efficiency}b). For each task and share, we draw five random subsets of the training studies, stratified by label and shared by all models, and we keep the validation and test sets whole. Each probe selects its regularization on the validation set and is refit as for the full probes. We leave a task out at a share when a subset has fewer than two studies of a class, which leaves 32 of 34 tasks in the private test set and 22 of 29 on MIMIC-IV-Echo at 1\%. We report the mean AUROC over tasks, averaged over the five subsets.

\textbf{Subgroup analysis.} We compare the linear-probe AUROC of the six models within patient subgroups: age, body mass index, scanner vendor and clinical center in the private test set, and age, sex, image quality, scanner model, body mass index and resting blood pressure on MIMIC-IV-Echo (Extended Data Fig.~\ref{fig:ed-subgroups}). On MIMIC-IV-Echo, age and sex come from the linked MIMIC-IV hospital records \cite{johnson2023mimiciv,PhysioNet-mimiciv-3.1}. A finding is evaluable in a subgroup that has at least 10 positive and 10 negative test studies for it. For each covariate, we average AUROC over the findings that are evaluable in every subgroup of at least 100 studies, and we do not show smaller subgroups. By clinical center, each center keeps its own evaluable findings. Differences between models are paired and use the bootstrap described in the Statistical analysis section.

\textbf{Segmentation.} We evaluate segmentation with encoder weights fixed on CAMUS, for the left ventricular cavity, left ventricular myocardium and left atrium, and on EchoNet-Dynamic and EchoNet-Pediatric, for the left ventricular cavity. Each model keeps its own input preprocessing, so the center crops of EchoPrime and EchoJEPA cut part of the CAMUS left atrium. On each model's features, we train a convolutional decoder for 60 epochs, with batch size 32 and weight decay $10^{-4}$, and select the learning rate and epoch by validation Dice similarity coefficient (Extended Data Table~\ref{tab:ed-cardiobench}d). We also train a separate decoder on CAMUS for the left ventricular cavity alone and test it on CAMUS, EchoNet-Dynamic and EchoNet-Pediatric (Extended Data Table~\ref{tab:ed-cardiobench}e). For EchoAlign, the features are the TimeSformer's space-time patch tokens at the annotated frame, arranged as a 14 $\times$ 14 map. These are the same tokens from which the anatomical queries extract their embeddings. The decoder uses three random seeds. We average Dice similarity coefficients per patient and estimate 95\% CIs and paired differences with 1,000 patient-level bootstrap replicates.

\subsection{Statistical analysis}\label{sec:statistics}

We report test AUROC for classification, RMSE in each target's units for measurement and QWK for severity grading \cite{cohen1968weighted}, and models are compared on the test studies that every model scores. The overall classification mean averages all 68 classification task-source pairs (34 in the private test set, 29 on MIMIC-IV-Echo and five on CardioBench) with equal weight, and the overall grading mean averages the 31 graded tasks (11 in the private test set and 20 on MIMIC-IV-Echo) in the same way. We estimate 95\% confidence intervals from the 2.5th and 97.5th percentiles of 1,000 cluster bootstrap replicates \cite{efron1993bootstrap,field2007bootstrapping} of the saved predictions, resampling studies in the private test set, patients in MIMIC-IV-Echo and, in CardioBench, patients for HMC-QU, SegRWMA and CAMUS and studies for the other datasets. Replicates in which a task has only one class are left out of that task and of every mean that includes it. Within each source, all models and tasks share the same replicates, so comparisons between models are paired. For each comparison between two models, the two-sided $P$ value is twice the smaller of two proportions of the paired bootstrap replicates: those in which the difference is at or below zero, and those in which it is at or above zero. We call a difference significant when $P < 0.05$, and $P$ values are not adjusted for multiple comparisons. We measure how continuous text scores rise with reported grade using Spearman's rank correlation \cite{spearman1904proof}. Differences are computed from unrounded estimates, so a reported difference can differ by 0.001 from the difference of the rounded values.

\textbf{Computing and software.} We pretrain and extract embeddings on AMD Instinct MI210 GPUs and fit probes on CPUs. The pipeline uses Python, PyTorch, DeepSpeed, scikit-learn, NumPy, SciPy, pandas and open\_clip.

\backmatter

% \bmhead{Reporting summary}

% Further information on research design is available in the Nature Portfolio Reporting Summary linked to this article.

\bmhead{Data availability}

The eight CardioBench datasets \cite{taratynova2025cardiobench} are available from their original sources: TMED-2 \cite{huang2022tmed}, CardiacNet \cite{yang2024cardiacnet},
SegRWMA \cite{liu2023enhance}, HMC-QU \cite{degerli2021early,degerli2023multiview}, EchoNet-Dynamic \cite{echonetdynamic2019}, EchoNet-LVH
\cite{echonetlvh2020}, EchoNet-Pediatric \cite{echonetpediatric2025} and CAMUS
\cite{leclerc2019deep}. The CardioBench code is also publicly available \cite{taratynova2025cardiobench}. MIMIC-IV-Echo and MIMIC-IV \cite{johnson2023mimiciv,PhysioNet-mimiciv-3.1} are available from PhysioNet \cite{pollard_physionet_2026} to credentialed users who
complete the required training and sign the data use agreement \cite{PhysioNet-mimic-iv-echo-1.0}.
The EchoAlign code repository contains the CardioBench splits and the scripts that derive our MIMIC-IV-Echo and CardioBench labels and splits from these releases.
The private dataset, provided by iCardio.ai, is not publicly available.

\bmhead{Code availability}

The EchoAlign model weights, with their principal-component projections, and evaluation code are available at \url{https://github.com/BioMedIA-MBZUAI/EchoAlign}.

% \bmhead{Author contributions} 
% \textbf{Research, code and analysis.} \textit{Core contributor:} A.A. \textit{Contributors:} D.T., L.A.M.\\
% \textbf{Data and clinical guidance.} \textit{Core contributors:} R.A.S., J.Z.S., J.P.\\
% \textbf{Supervision.} \textit{Core contributors:} N.S., M.Y. \textit{Contributor:} K.H.M.-H.\\
% \textbf{Paper writing.} \textit{Core contributor:} A.A. \textit{Contributors:} all other authors.

% \bmhead{Competing interests}

% R.A.S., J.Z.S. and J.P. are employees of iCardio.ai. The other authors declare no competing interests.

% \bmhead{Ethics approval}

% This study used only retrospective echocardiographic data. The private dataset, provided by iCardio.ai, was de-identified before analysis and fell outside the scope of institutional review board oversight. MIMIC-IV-Echo was used under its data use agreement, with credentialed access through PhysioNet \cite{PhysioNet-mimic-iv-echo-1.0}. The CardioBench datasets are publicly available \cite{taratynova2025cardiobench}. No prospective data collection or human participant intervention was performed.

\begin{appendices}
\renewcommand{\tablename}{Extended Data Table}
\renewcommand{\theHtable}{ed.\arabic{table}}

\clearpage
\section{Extended Data}\label{secA1}
\makeatletter\setlength{\@fptop}{0pt}\setlength{\@fpbot}{0pt plus 1fil}\makeatother
\setcounter{figure}{0}
\renewcommand{\figurename}{Extended Data Fig.}
\renewcommand{\theHfigure}{ed.\arabic{figure}}

\begin{figure}[!h]
\vspace{-\intextsep}
\makebox[\textwidth][c]{\begin{minipage}{\textwidth}
\centering
\begin{minipage}{130mm}
\includegraphics[width=\linewidth]{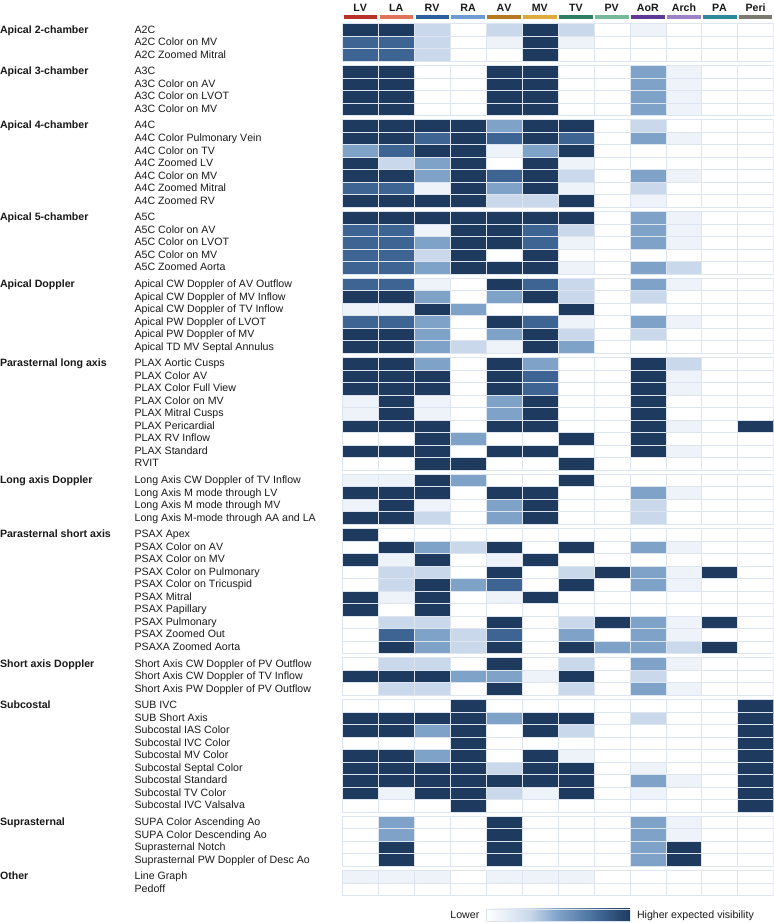}
\end{minipage}
\vspace{2mm}
\caption{\textbf{Visibility matrix.}
Expected visibility of each of the 12 cardiac structures (columns) in each of the 66 views (rows). We set the weights using American Society of Echocardiography guidelines as the primary reference \cite{mitchell2019comprehensive}. Darker cells mark higher expected visibility, and white cells mark none. LV, left ventricle; LA, left atrium; RV, right ventricle; RA, right atrium; AV, aortic valve; MV, mitral valve; TV, tricuspid valve; PV, pulmonic valve; AoR, aortic root; Arch, aortic arch; PA, pulmonary artery; Peri, pericardium; A2C to A5C, apical two-chamber to five-chamber; PLAX, parasternal long axis; PSAX, parasternal short axis; RVIT, right ventricular inflow tract; SUB, subcostal; SUPA, suprasternal.}\label{fig:ed-visibility}
\end{minipage}}
\end{figure}

\begin{figure}[p]
\makebox[\textwidth][c]{\begin{minipage}{\textwidth}
\centering
\begin{minipage}{130mm}
\begin{minipage}[t]{\textwidth}
\panel{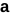}\\[0.3ex]
\includegraphics[width=\linewidth]{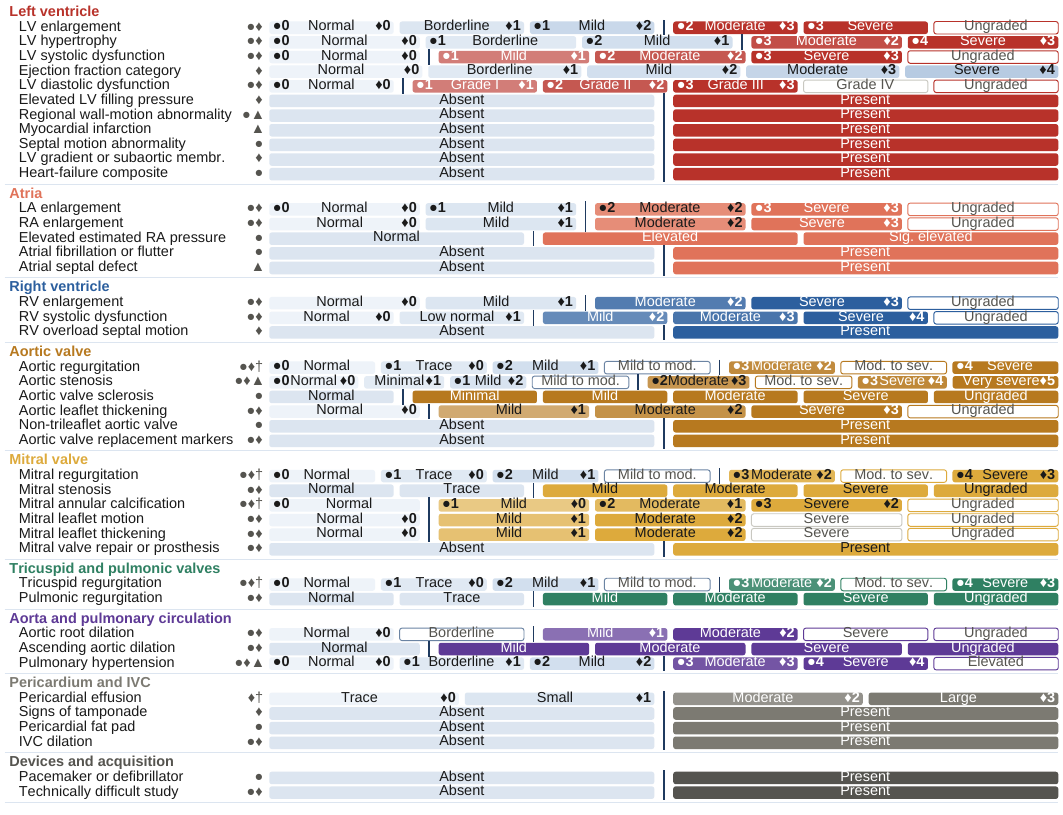}
\end{minipage}\\[1.2ex]
\begin{minipage}[t]{\textwidth}
\panel{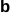}\\[0.3ex]
\includegraphics[width=\linewidth]{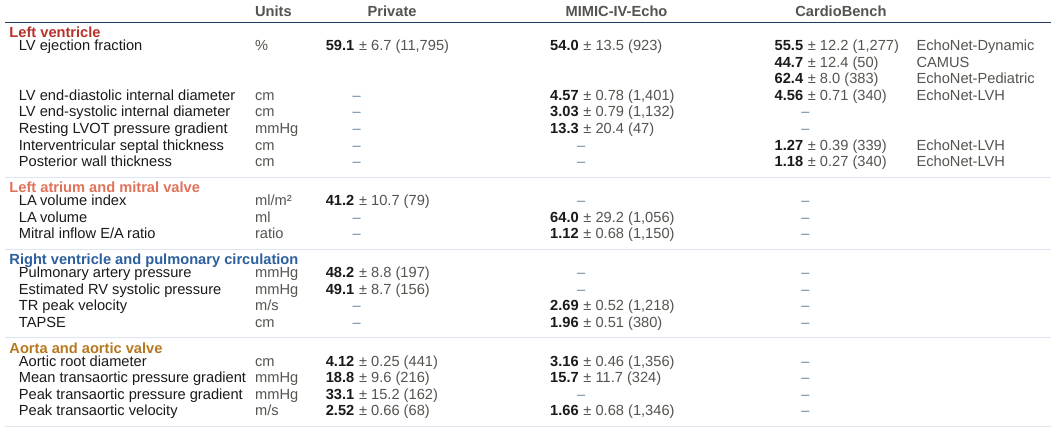}
\end{minipage}
\end{minipage}
\vspace{2mm}
\caption{\textbf{Label definitions and measurement targets.}
\textbf{a}, Grades of each finding for classification and severity grading, lowest to highest. Pale blue is negative and color positive; the rule marks the boundary. Numbers give the ordinal grade (0 lowest) in the private dataset (\textbullet, left) and MIMIC-IV-Echo ($\blacklozenge$, right); outlined grades are binary only. MIMIC-IV-Echo records the findings marked $\dagger$ only when present, so we grade them from their lowest recorded grade. Extended Data Table~\ref{tab:ed-regression-grading}b shows which findings are graded in each dataset. The markers after each finding name show the datasets in which it is a task; $\blacktriangle$ is CardioBench.
\textbf{b}, Test-label mean (bold), with standard deviation and test count. Dashes mark targets not defined in that dataset. LV, left ventricle; LA, left atrium; RV, right ventricle; RA, right atrium; IVC, inferior vena cava; LVOT, left ventricular outflow tract; TR, tricuspid regurgitation; TAPSE, tricuspid annular plane systolic excursion; E/A, ratio of early to late mitral inflow velocity; membr., membrane.}\label{fig:ed-labels}
\end{minipage}}
\end{figure}

\begin{figure}[p]
\makebox[\textwidth][c]{\begin{minipage}{\textwidth}
\centering
\begin{minipage}{130mm}
\begin{minipage}[t]{\textwidth}
\panel{a}\\[0.3ex]
\includegraphics[width=\linewidth]{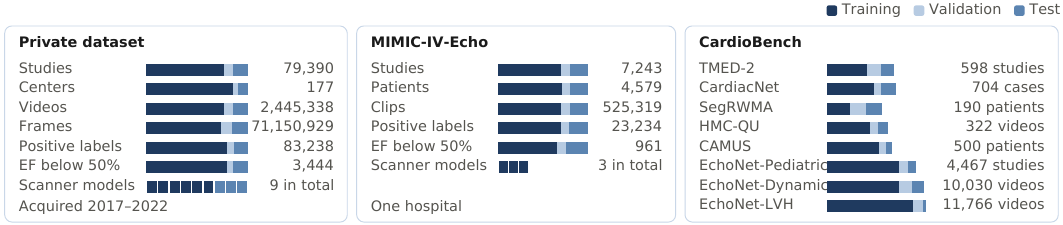}
\end{minipage}\\[0.6ex]
\begin{minipage}[t]{0.4578\textwidth}
\panel{b}\\[0.3ex]
\includegraphics[width=\linewidth]{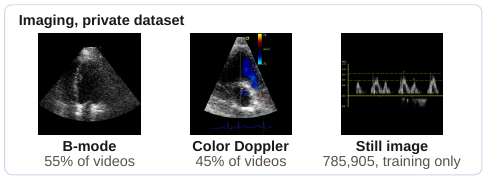}
\end{minipage}%
\begin{minipage}[t]{0.5422\textwidth}
\panel{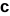}\\[0.3ex]
\includegraphics[width=\linewidth]{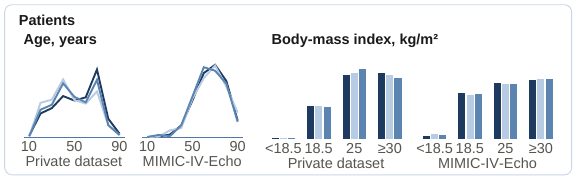}
\end{minipage}\\[0.6ex]
\begin{minipage}[t]{\textwidth}
\panel{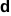}\\[0.3ex]
\includegraphics[width=\linewidth]{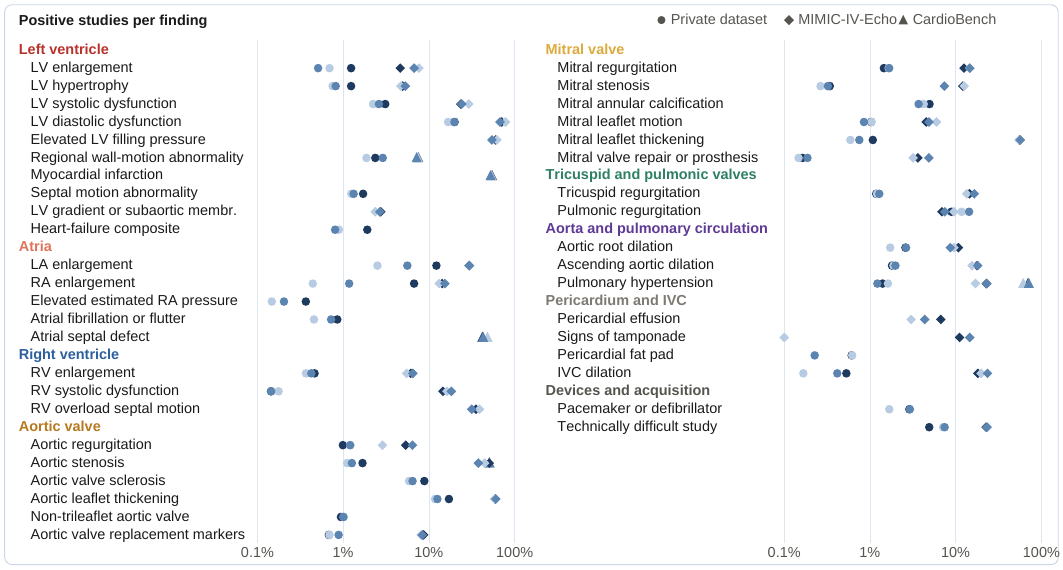}
\end{minipage}
\end{minipage}
\vspace{2mm}
\caption{\textbf{Evaluation datasets.}
\textbf{a}, Size of each dataset. Each bar divides a count into training, validation and test sets, and the number gives the total. The private dataset is divided by center, MIMIC-IV-Echo by patient and CardioBench by its released splits, in each dataset's prediction unit. Positive labels sum positive studies over all classification tasks. Each scanner cell is one scanner model: navy if it occurs in training, mid blue if it occurs only in validation or test.
\textbf{b}, Example B-mode, color Doppler and still frames from the private dataset, with each type's share of videos. Still images occur in the training set only.
\textbf{c}, Age by decade (lines) and body-mass index category (bars) as a percentage of studies in each set.
\textbf{d}, Percentage of positive studies per finding in each set, on a logarithmic scale. Marker shape shows the dataset and color the set. Extended Data Fig.~\ref{fig:ed-labels} defines a positive study for each finding. EF, ejection fraction; LV, left ventricle; LA, left atrium; RV, right ventricle; RA, right atrium; IVC, inferior vena cava; membr., membrane.}\label{fig:ed-data}
\end{minipage}}
\end{figure}

\begin{figure}[p]
\makebox[\textwidth][c]{\begin{minipage}{\textwidth}
\centering
\begin{minipage}{130mm}
\includegraphics[width=\linewidth]{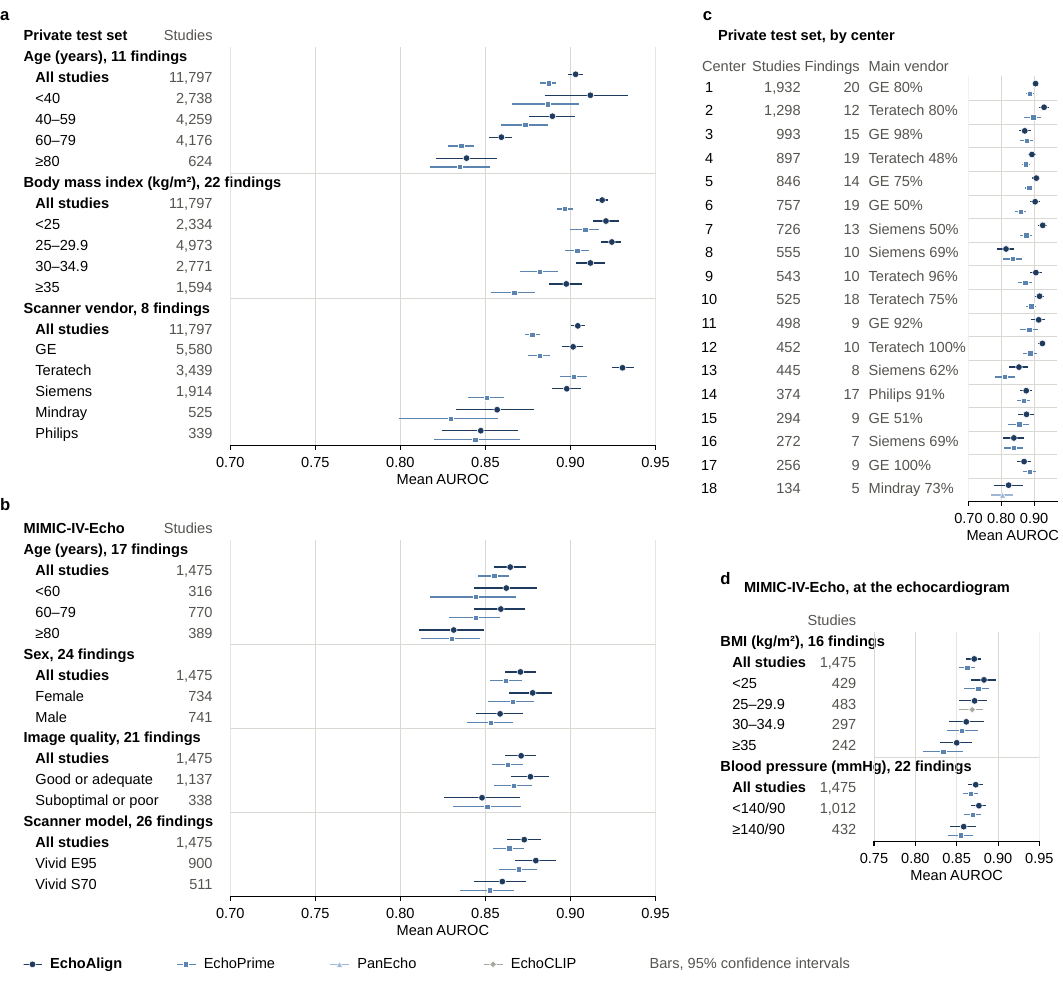}
\end{minipage}
\vspace{2mm}
\caption{\textbf{Classification performance in patient subgroups.}
Mean AUROC of linear probes for EchoAlign and, on each row, the best of the other five models, within each subgroup, over the findings evaluable in every subgroup of a covariate.
\textbf{a}, Private test set by age, body mass index and scanner vendor.
\textbf{b}, MIMIC-IV-Echo by age, sex, image quality and scanner model.
\textbf{c}, Private test set by center, over the findings evaluable at each center. Main vendor is the most frequent vendor among the center's studies.
\textbf{d}, MIMIC-IV-Echo by body mass index and resting blood pressure recorded at the echocardiogram. Blood pressure is a single reading and does not define hypertension.
MIMIC-IV-Echo panels cover the test studies with at least one task label. Underweight studies are included in the lowest body-mass-index group. Levels with fewer than 100 studies (Vivid E90, 64 studies) are not shown. Bars show 95\% bootstrap confidence intervals. BMI, body mass index.}\label{fig:ed-subgroups}
\end{minipage}}
\end{figure}

\begin{figure}[p]
\makebox[\textwidth][c]{\begin{minipage}{\textwidth}
\centering
\begin{minipage}{130mm}
\begin{minipage}[t]{0.49\textwidth}
\panel{a}\\[0.3ex]
\includegraphics[width=\linewidth]{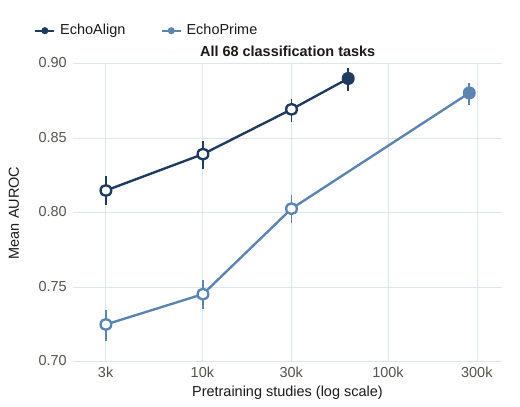}
\end{minipage}\hfill
\begin{minipage}[t]{0.49\textwidth}
\panel{b}\\[0.3ex]
\includegraphics[width=\linewidth]{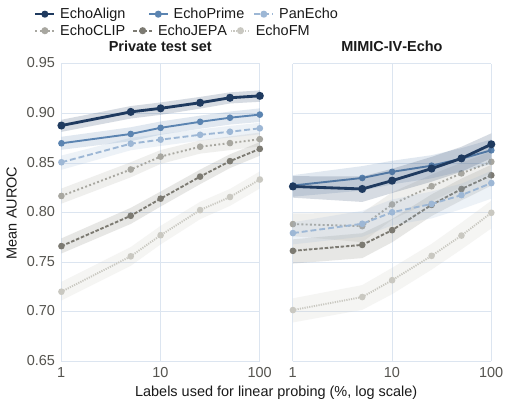}
\end{minipage}
\end{minipage}
\vspace{2mm}
\caption{\textbf{Pretraining and label efficiency.}
\textbf{a}, Mean AUROC over 68 classification tasks against the number of pretraining studies, for the EchoAlign and EchoPrime training recipes on the same nested subsets (hollow circles). Filled circles are the released models trained on their full data, 60,782 studies for EchoAlign and 272,256 for EchoPrime \cite{vukadinovic2026comprehensive}; EchoPrime's released model was trained on a different archive. Filled circles use the readout protocol of this analysis. Bars show 95\% bootstrap confidence intervals. At 30,000 studies, EchoAlign's mean AUROC exceeds that of the EchoPrime recipe by 0.067 ($P < 0.01$).
\textbf{b}, Mean AUROC against the share of training labels used for linear probing, for the six released models (filled circles), in the private test set (left) and on MIMIC-IV-Echo (right). Lines show the mean over tasks across five random label subsets, and bands the 95\% bootstrap confidence interval. At 1\% of the labels we drop tasks with fewer than two studies of a class in a subset, leaving 32 of 34 tasks in the private test set and 22 of 29 on MIMIC-IV-Echo; all tasks are included from 5\%.}\label{fig:ed-efficiency}
\end{minipage}}
\end{figure}

\begin{figure}[p]
\makebox[\textwidth][c]{\begin{minipage}{\textwidth}
\centering
\includegraphics[width=\linewidth]{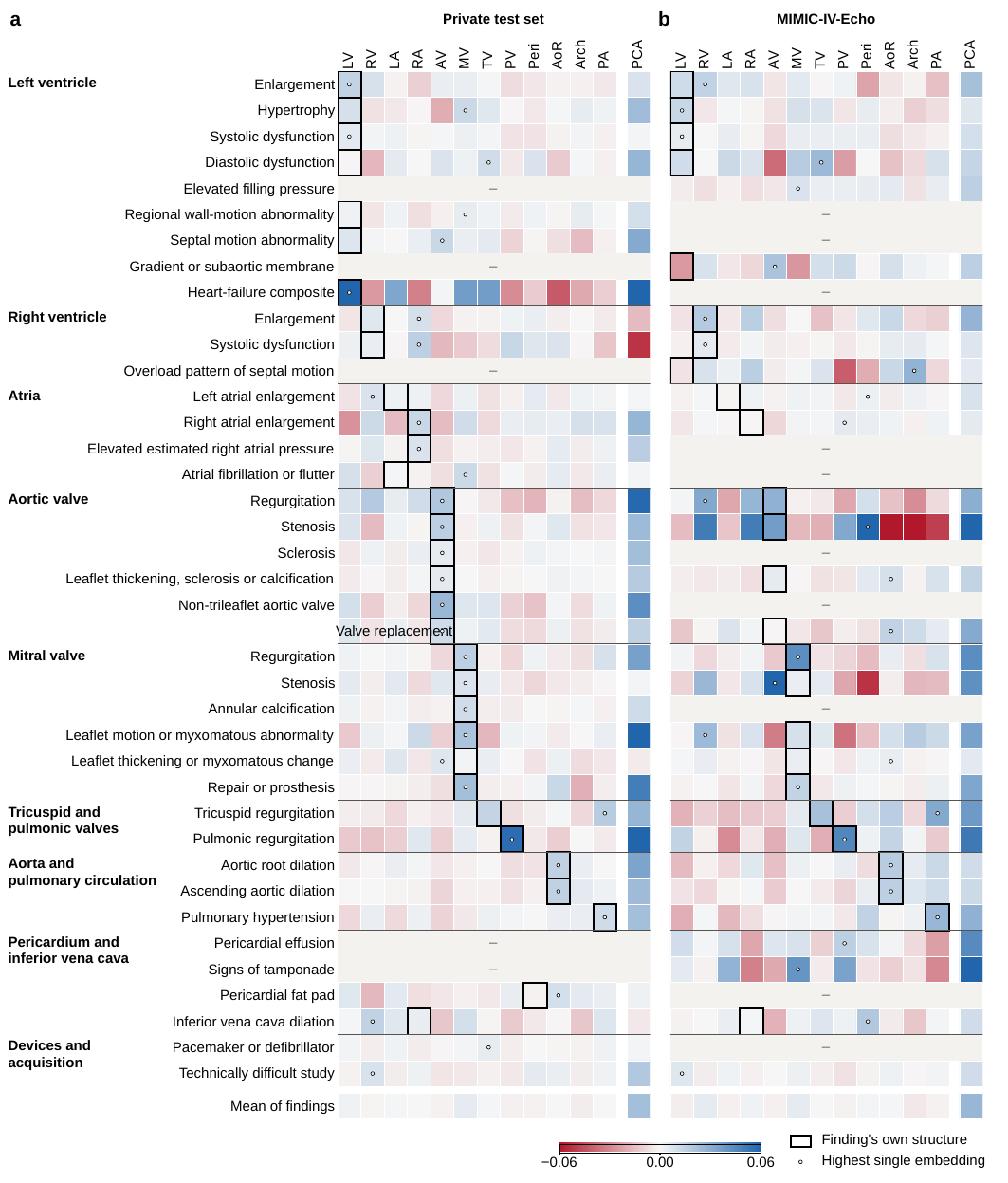}
\caption{\textbf{Each finding read from each study embedding.} AUROC of a linear probe on each of the 12 anatomical study embeddings and on the study-level embedding, which combines the global and the 12 anatomical study embeddings with principal component analysis (PCA) (columns) for each finding (rows), grouped as in Extended Data Table~\ref{tab:ed-classification}, in the private test set (\textbf{a}) and on MIMIC-IV-Echo (\textbf{b}). Values are AUROC above or below the mean of the 12 embeddings for that finding, and the bottom row averages the findings. Outlines mark the embedding of a finding's own cardiac structure, where one is assigned, and dots mark the highest of the 12 embeddings for a finding. Dashes mark findings not included for that test set. LV, left ventricle; RV, right ventricle; LA, left atrium; RA, right atrium; AV, aortic valve; MV, mitral valve; TV, tricuspid valve; PV, pulmonic valve; Peri, pericardium; AoR, aortic root; Arch, aortic arch; PA, pulmonary artery.}\label{fig:ed-token-map}
\end{minipage}}
\end{figure}

\gdef\floatpagefraction{1.01}
\begin{table}[p]
\makebox[\textwidth][c]{\begin{minipage}{180mm}
\caption{\textbf{Classification without a fitted probe in the private test set and MIMIC-IV-Echo.} AUROC of EchoAlign, EchoPrime and EchoCLIP scored with text prompts, and of PanEcho's task heads, which are trained on labels; values are shown $\pm$ half the width of the 95\% bootstrap confidence interval, in gray. Where the value plus half the width would pass 1, the cell gives the interval limits. Bold marks the highest value, and in mean rows the highest among models that score every task; dashes mark tasks absent from a test set or without a PanEcho head. $^{*}$EchoPrime or EchoCLIP scored with EchoAlign's prompts, because its released prompts or report retrieval do not cover the task. $^{\ddagger}$Mean over the tasks PanEcho covers: 18 in each source.}\label{tab:ed-from-text}
\begin{tabular}{@{}c@{}}\makebox[180mm][c]{\includegraphics[width=155.4mm]{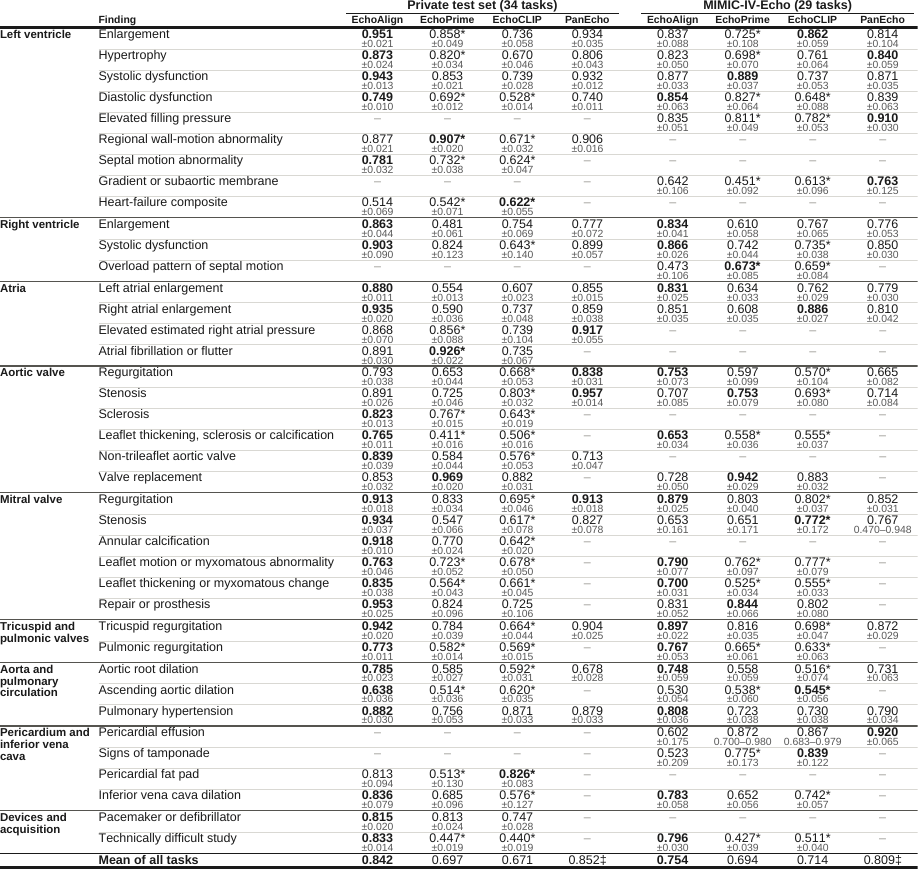}}\end{tabular}
\end{minipage}}
\end{table}
\begin{table}[p]
\makebox[\textwidth][c]{\begin{minipage}{180mm}
\caption{\textbf{Performance in CardioBench and segmentation.} \textbf{a}, AUROC with text prompts, and of PanEcho's task heads. \textbf{b}, AUROC of linear probes. \textbf{c}, RMSE of linear probes. \textbf{d}, Dice similarity coefficient of a decoder trained and tested on the same dataset. \textbf{e}, Left ventricular cavity with a separate decoder, trained on CAMUS for this structure alone and tested on each dataset. Bold marks the best value, and values show $\pm$ half the width of the 95\% bootstrap confidence interval. Dashes mark tasks without a PanEcho head. $^{\dagger}$EchoPrime scores TMED-2 from still images repeated into clips. In panel \textbf{a}, EchoPrime and EchoCLIP use EchoAlign's or CardioBench's prompts on tasks their own methods do not cover; $^{\ddagger}$mean over the three tasks PanEcho covers.}\label{tab:ed-cardiobench}
\begin{tabular}{@{}c@{}}\makebox[180mm][c]{\includegraphics[width=140.5mm]{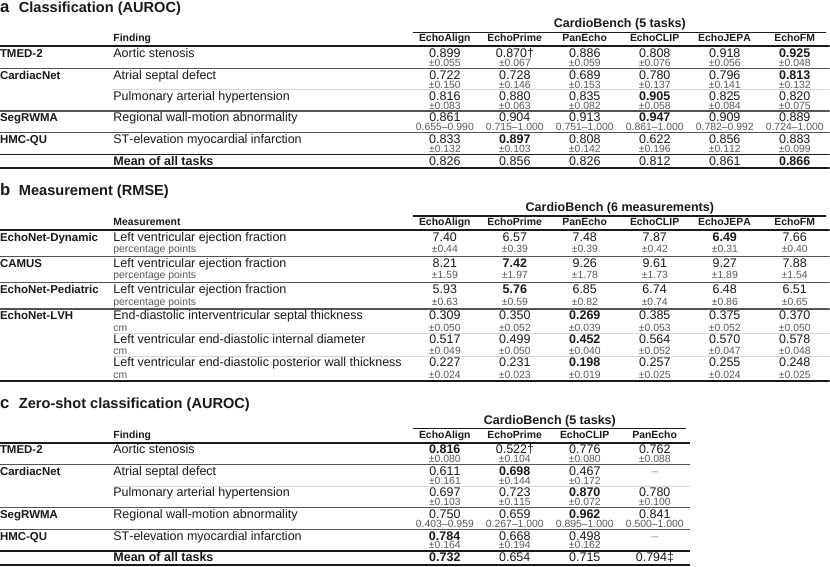}}\\[1mm]\makebox[180mm][c]{\includegraphics[width=111.74mm]{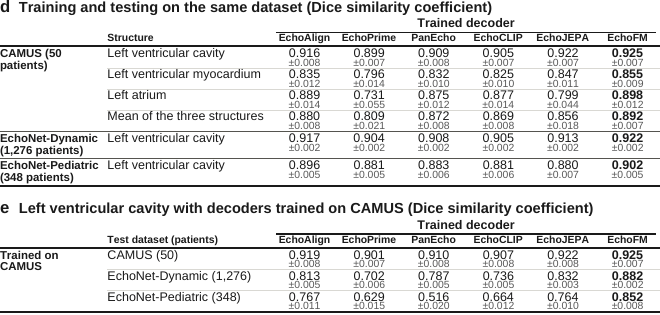}}\end{tabular}
\end{minipage}}
\end{table}
\begin{table}[p]
\makebox[\textwidth][c]{\begin{minipage}{179.9mm}
\caption{\textbf{Classification performance in the private test set and MIMIC-IV-Echo.} AUROC of linear probes on pretrained embeddings; values are shown $\pm$ half the width of the 95\% bootstrap confidence interval, in gray. Where the value plus half the width would pass 1, the cell gives the interval limits. Bold marks the best linear probe; dashes mark tasks absent from a test set. Extended Data Fig.~\ref{fig:ed-labels} defines each task, and Extended Data Fig.~\ref{fig:ed-data}d gives its percentage of positive studies.}\label{tab:ed-classification}
\begin{tabular}{@{}c@{}}\includegraphics[width=180mm]{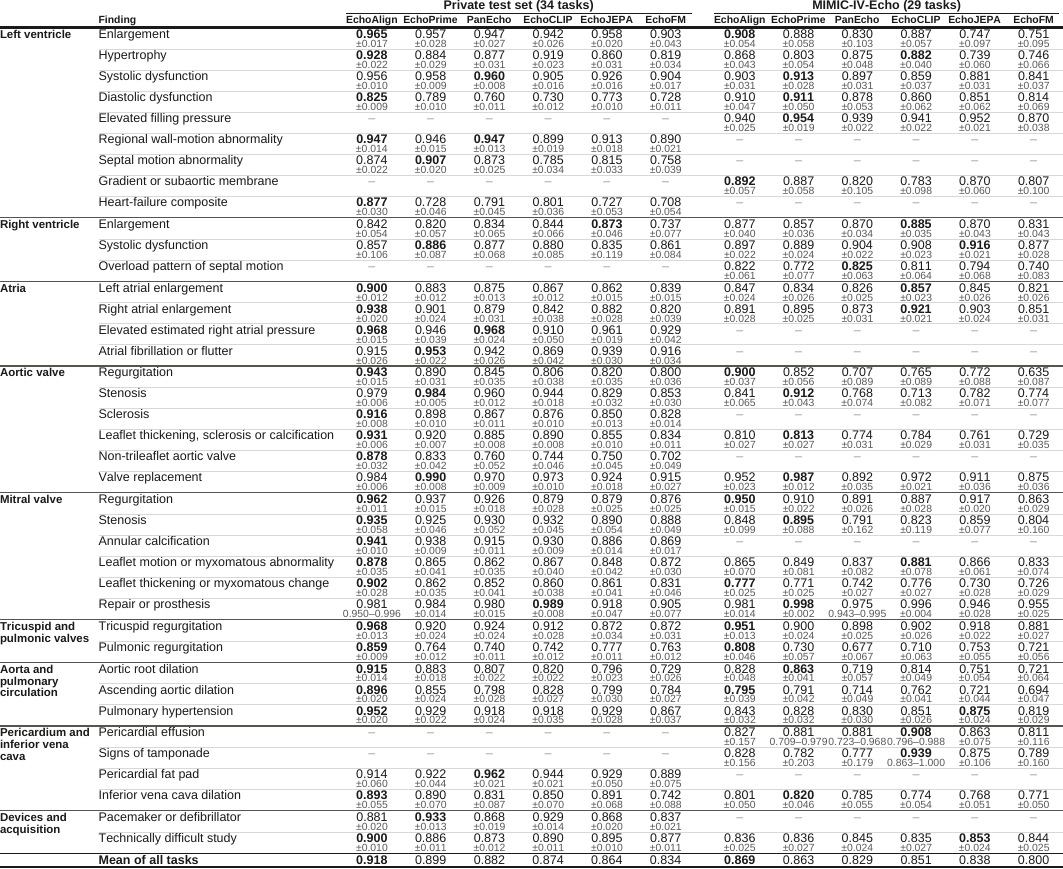}\end{tabular}
\end{minipage}}
\end{table}
\begin{table}[p]
\makebox[\textwidth][c]{\begin{minipage}{180mm}
\caption{\textbf{Measurements and severity grading.} \textbf{a}, RMSE of linear probes in the private test set and MIMIC-IV-Echo; bold marks the lowest value. \textbf{b}, QWK of ordinal probes in the private test set and MIMIC-IV-Echo; bold marks the highest value, and dashes mark findings not graded in that test set. Each finding keeps its native grade scale in each test set, with the top two grades of a MIMIC-IV-Echo finding merged when a grade is missing from a set, as described in the Methods section. $^\dagger$MIMIC-IV-Echo records these findings only when present, so we grade them from their lowest recorded grade; its mean includes them. Values are shown $\pm$ half the width of the 95\% bootstrap confidence interval, in gray. RMSE, root-mean-square error; QWK, quadratic weighted kappa; TAPSE, tricuspid annular plane systolic excursion.}\label{tab:ed-regression-grading}
\begin{tabular}{@{}c@{}}\includegraphics[width=180mm]{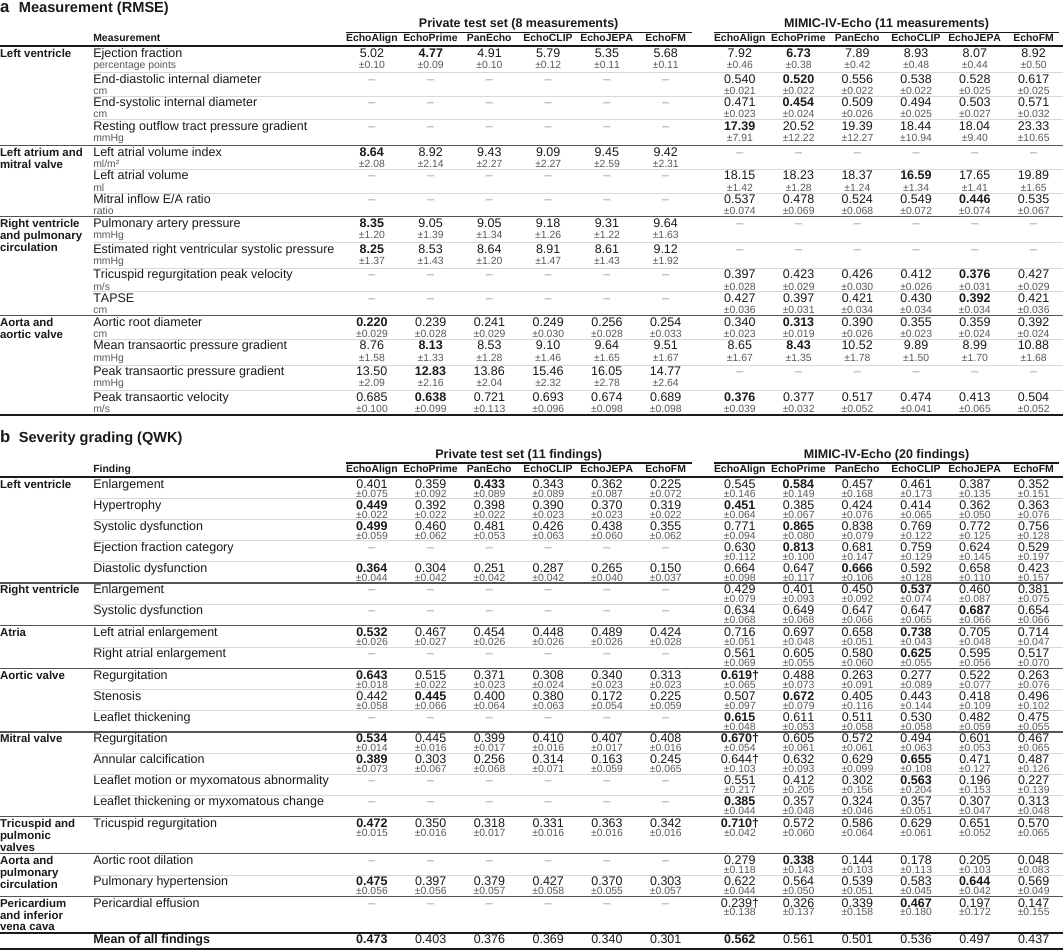}\end{tabular}
\end{minipage}}
\end{table}
\end{appendices}

\end{document}